\documentclass[sigconf]{acmart}
\AtBeginDocument{%
	}

\setcopyright{acmlicensed}
\copyrightyear{2018}
\acmYear{2018}
\acmDOI{XXXXXXX.XXXXXXX}
\acmConference[Conference acronym 'XX]{Make sure to enter the correct
	conference title from your rights confirmation email}{June 03--05,
	2018}{Woodstock, NY}

\acmISBN{978-1-4503-XXXX-X/2018/06}

\usepackage{enumitem}
\usepackage{booktabs}
\usepackage{longtable}
\usepackage{multirow}
\usepackage[table]{xcolor}
\usepackage{graphicx}
\usepackage{algorithm}
\usepackage{algpseudocode}
\usepackage{amsmath}
\usepackage{subcaption}
\graphicspath{{figures/}}

\definecolor{lightgray}{gray}{0.92} % 0:black -> 1: white
\definecolor{darkgray}{gray}{0.85}
\begin{document}

	\title{uLEAD-TabPFN: Uncertainty-aware Dependency-based Anomaly Detection with TabPFN}
	
%	\author{Anonymous For Review}
%	\affiliation{%
%		\institution{Anonymous Institution(s)}
%		\city{Anonymous}
%		\state{Anonymous}
%		\country{Anonymous}
%	}
%	\email{anonymous@anonymous.edu}
	
	\author{Sha Lu}
	\author{Jixue Liu}
	\author{Stefan Peters}
	\affiliation{%
		\institution{Adelaide University}
		\city{}
		\state{South Australia}
		\postcode{5000}
		\country{Australia}
	}
	\email{sha.lu@adelaide.edu.au}
	% \email{{sha.lu,jixue.liu,stefan.peters,}@adelaide.edu.au}
	
	\author{Thuc Duy Le}
	\author{Lin Liu}
	\author{Jiuyong Li}
	\affiliation{%
		\institution{Adelaide University}
		\city{}
		\state{South Australia}
		\postcode{5000}
		\country{Australia}
	}
	% \email{{thuc.le,lin.liu,jiuyong.li}@adelaide.edu.au}
	
	\author{Yongzheng Xie}
	\affiliation{%
		\institution{Queen's University Belfast}
		\city{}
		\state{Northern Ireland}
		\postcode{BT7 1NN}
		\country{United Kingdom}
	}
	\email{c.xie@qub.ac.uk}
	
	% after all \author / \affiliation blocks
	\authorsaddresses{
		Correspondence to: Sha Lu (sha.lu@adelaide.edu.au).
	}

	\begin{abstract}
		Anomaly detection in tabular data is challenging due to high dimensionality, complex feature dependencies, and heterogeneous noise. 
		Many existing methods rely on proximity-based cues and may miss anomalies caused by violations of complex feature dependencies.		
		Dependency-based anomaly detection provides a principled alternative by identifying anomalies as violations of dependencies among features.
		However, existing methods often struggle to model such dependencies robustly and to scale to high-dimensional data with complex dependency structures.
		To address these challenges, we propose uLEAD-TabPFN, a dependency-based anomaly detection framework built on Prior-Data Fitted Networks (PFNs).
		uLEAD-TabPFN identifies anomalies as violations of conditional dependencies in a learned latent space, leveraging frozen PFNs for dependency estimation.
		Combined with uncertainty-aware scoring, the proposed framework enables robust and scalable anomaly detection.		
		Experiments on 57 tabular datasets from ADBench show that uLEAD-TabPFN achieves particularly strong performance in medium- and high-dimensional settings, where it attains the top average rank. On high-dimensional datasets, uLEAD-TabPFN improves the average ROC-AUC by nearly 20\% over the average baseline and by approximately 2.8\% over the best-performing baseline, while maintaining overall superior performance compared to state-of-the-art methods.
		Further analysis shows that uLEAD-TabPFN provides complementary anomaly detection capability, achieving strong performance on datasets where many existing methods struggle.		
	\end{abstract}

	
%	\received{20 February 2007}
%	\received[revised]{12 March 2009}
%	\received[accepted]{5 June 2009}
	
	\maketitle

	\section{Introduction}
	\label{sec:introduction}	
	Anomaly detection is a core problem in data science with critical applications in areas such as cybersecurity, manufacturing, finance, healthcare, and environmental monitoring~\cite{chandola2009anomaly,pang2021deep,aggarwal2016introduction}.
	Despite substantial progress, detecting anomalies in tabular data remains challenging due to mixed feature types, high dimensionality, and complex dependencies among features~\cite{thudumu2020highdim,chavent2021mixed,zoppi2024tabularIDS}.	

	Most existing approaches for tabular anomaly detection are proximity-based, identifying anomalies based on proximity relationships among samples~\cite{chandola2009anomaly,ruff2021unifying,ramaswamy2000knn,breunig2000lof,liu2008iforest}.
	They are inherently limited in capturing anomalies arising from violations of complex inter-feature dependencies.
	In addition, proximity-based criteria are known to degrade in high-dimensional settings due to the curse of dimensionality~\cite{thudumu2020highdim,chavent2021mixed,zoppi2024tabularIDS}, further limiting their effectiveness.

	Dependency-based anomaly detection offers a principled alternative by explicitly modeling relationships among variables and identifying anomalies as violations of dependency structure~\cite{aggarwal2016introduction,pang2021deep}.
	Prior work has demonstrated that such methods are effective in detecting subtle anomalies that may not manifest as proximity-based deviations~\cite{lu2025dependency,tsai2025anollm,xie2021logdp,buck2023admire}.
	
	However, existing dependency-based approaches face several fundamental challenges.
	First, many methods model conditional dependencies directly in the original feature space~\cite{lu2025dependency,lu2020lopad,li2023quantile,Hirth2023ncsnad,homayouni2024anomaly,yin2024mcm}, which limits both their robustness and computational efficiency in high-dimensional settings.
	Second, many existing approaches~\cite{feng2024pars,li2025trustworthy,zhao2025tadgp,tsai2025anollm} rely on rigid, task-specific models, limiting robustness and scalability as feature dimensionality and dependency complexity increase, especially when training data are limited.
	Third, existing approaches~\cite{lu2025dependency,huynh2024dagnosis,li2023quantile,elbahaay2023unsupervised,Hirth2023ncsnad} typically rely on dependency deviations as anomaly evidence, while largely overlooking uncertainty in dependency estimation.
	As a result, anomaly scoring can be unreliable in heterogeneous or noisy settings.

	Motivated by the above challenges, we propose uLEAD-TabPFN, an uncertainty-aware dependency-based anomaly detection framework designed to enable robust and scalable modeling of conditional dependencies in high-dimensional tabular data.
	To facilitate reliable dependency modeling, uLEAD-TabPFN learns a low-dimensional latent representation aligned with dependency evaluation.
	This representation both alleviates challenges associated with high-dimensional feature spaces and also makes dependency relationships more explicit in the latent space.
	To overcome the fragility of task-specific dependency estimators, uLEAD-TabPFN leverages Prior-Data Fitted Networks (PFNs)~\cite{hollmann2025tabpfn} as frozen conditional predictors.
	By exploiting their rich pre-trained priors, PFNs enable expressive dependency modeling via in-context inference without requiring dataset-specific model training.
	Finally, to account for variability in normal dependency relationships, uLEAD-TabPFN incorporates uncertainty-aware scoring, enabling dependency
	deviations to be evaluated relative to their estimated uncertainty.
	
	Together, these design choices enable uLEAD-TabPFN to improve the robustness, scalability, and reliability of dependency-based anomaly detection, even in challenging high-dimensional settings.
	
	Extensive experiments on diverse tabular datasets from ADBench~\cite{han2022adbench}, a widely used benchmark for systematic evaluation of tabular anomaly detection methods, show that uLEAD-TabPFN attains the top average rank on medium- and high-dimensional datasets, while achieving competitive or superior performance compared to existing approaches overall.
	Furthermore, uLEAD-TabPFN exhibits performance patterns that differ from those of the 26 baselines, suggesting that it captures complementary anomaly evidence rather than providing incremental improvements over existing approaches.

	Our main contributions are summarized as follows:
	\begin{itemize}
		\item We propose uLEAD-TabPFN, a dependency-based anomaly detection framework that detects anomalies as violations of conditional dependencies by combining representation learning, pre-trained foundation models, and uncertainty-aware dependency evaluation.
		
		\item We demonstrate that reliable conditional dependency estimation can be achieved by leveraging pre-trained PFNs as frozen predictors, avoiding dataset-specific dependency model training while improving the robustness and efficiency of dependency-based anomaly detection.
		
		\item We show that uLEAD-TabPFN captures anomaly evidence that is complementary to existing approaches, exhibiting detection behavior that differs from current state-of-the-art (SOTA) methods rather than constituting an incremental refinement.
		
		\item Through extensive experiments on diverse tabular benchmarks, we demonstrate that uLEAD-TabPFN achieves particularly strong performance on medium- and high-dimensional datasets, with consistently competitive overall performance.		
	\end{itemize}
	
	\section{Related Work}
	\label{sec:related_work}
	
	\paragraph{Proximity-based Tabular Anomaly Detection.}
	Early work on tabular anomaly detection primarily focuses on proximity-based methods, including distance- and density-based techniques, which identify anomalies as isolated points or instances in low-density regions of the feature space~\cite{chandola2009anomaly,aggarwal2016introduction,liu2008iforest,ramaswamy2000knn,breunig2000lof}.
	These methods rely on meaningful neighborhood structure and often degrade in high-dimensional settings, where distances become less informative and local neighborhoods lose discriminative power~\cite{ruff2021unifying,aggarwal2016introduction}.
		
	\paragraph{PFNs for Tabular Anomaly Detection.}
	Recent advances in foundation models have enabled in-context and zero-shot learning for tabular data~\cite{hollmann2025tabpfn,ren2025foundation}. 
	PFNs are pre-trained on large collections of synthetic datasets and perform inference by conditioning on a small context set~\cite{muller2022pfn,hollmann2025tabpfn}. 
	These models achieve strong performance in supervised tabular learning tasks without dataset-specific fine-tuning~\cite{feuer2023scaling,wang2025boostpfn}. 
	However, they are primarily designed for supervised settings and are not directly applicable to unsupervised or semi-supervised anomaly detection. 
	Only recently have PFN-based models begun to be explored for anomaly detection in zero-shot settings~\cite{fomo0d2024zeroshoot}.
	
	\paragraph{Dependency-based Anomaly Detection.}
	Dependency-based approaches address the limitations of proximity-based approaches by modeling relationships among variables and identifying anomalies as violations of learned dependency structures~\cite{lu2020lopad, lu2025dependency, buck2023admire, xie2021logdp}. Approaches such as DepAD~\cite{lu2025dependency} and ADMIRE~\cite{buck2023admire} demonstrate that conditional dependencies provide a strong detection signal beyond marginal outliers, while structural methods like DAGnosis~\cite{huynh2024dagnosis} and PARs~\cite{feng2024pars} explicitly localize inconsistencies within causal or logical frameworks. Furthermore, recent work has explored capturing these dependencies through masked cell modeling~\cite{yin2024mcm} and attention-based transformer architectures~\cite{homayouni2024anomaly} to better represent complex feature interactions.

	\section{Methodology}
	\label{sec:methodology}
	
	\subsection{Problem Setting and Notation}
	Let $\mathcal{X} \subset \mathbb{R}^d$ denote a $d$-dimensional tabular feature space.
	We consider a semi-supervised anomaly detection setting, where the training data
	$\mathcal{X}_{\mathrm{train}} = \{\mathbf{x}_i\}_{i=1}^{N}$
	consists exclusively of normal instances drawn from an unknown data-generating distribution.
	
	The goal is to learn an anomaly scoring function $s: \mathcal{X} \rightarrow \mathbb{R}$ such that, for a test sample $\mathbf{x} \in \mathcal{X}_{\mathrm{test}}$, larger values of $s(\mathbf{x})$ indicate a higher likelihood that $\mathbf{x}$ is anomalous. 
	The learned scores can be used either for ranking samples by abnormality or for threshold-based anomaly detection, depending on the application.
	
	In uLEAD-TabPFN, we introduce a latent representation space $\mathbb{R}^p$, where each input sample $\mathbf{x}_i \in \mathbb{R}^d$ is mapped to a latent vector $\mathbf{z}_i \in \mathbb{R}^p$.
	We denote the $j$-th latent component of sample $i$ by $z_{ij}$, and use $\mathbf{z}_{i,-j}$ to represent the collection of all remaining latent components excluding $z_{ij}$.

	% -------- uLEAD-TabPFN Framework ---------
	\begin{figure*}[htbp]
		\centering
		\includegraphics[width=1\textwidth]{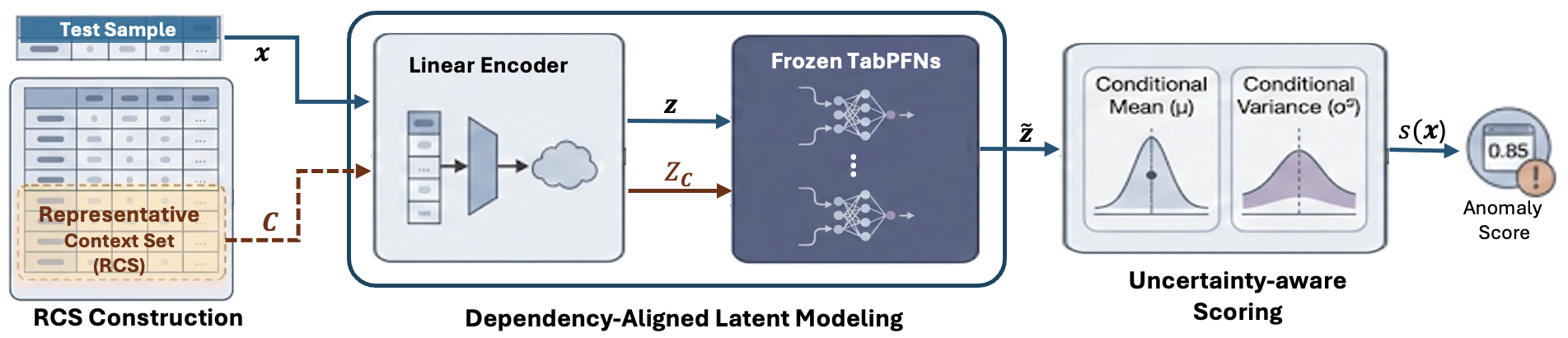}
		\caption{
			The uLEAD-TabPFN framework consists of three components: Representative Context Set (RCS) construction, dependency-aligned latent modeling, and uncertainty-aware scoring. 
			An RCS is constructed during training and used as a fixed in-context reference set. 
			During inference, both test samples and context samples are mapped into a dependency-aligned latent space using a lightweight encoder. 
			A frozen TabPFN predicts the conditional mean of each latent dimension by conditioning on the remaining dimensions and the RCS. 
			Anomaly scores are computed by aggregating dependency deviations normalized by learned variance estimates.
		}		
		\label{fig:framework}
	\end{figure*}

	\subsection{Overview of uLEAD-TabPFN}	
	Dependency-based anomaly detection~\cite{aggarwal2016introduction,lu2025dependency} characterizes anomalies as violations of conditional relationships among variables, assessing whether the value of a target variable is consistent with its conditional distribution given other variables.
	In practice, this paradigm is often instantiated by evaluating variables in a dimension-wise manner, conditioning each variable on the others, and measuring deviations from the learned conditional dependency structure.
	
	In this work, we adopt this paradigm and instantiate it through an in-context learning framework built on TabPFN~\cite{hollmann2025tabpfn}.
	TabPFN is a pre-trained probabilistic model trained on large-scale synthetic tabular data, enabling it to capture a wide range of data distributions.	
	It operates under an in-context learning paradigm, where predictions for a target variable are obtained by conditioning on a small set of reference samples, without requiring task-specific retraining.
	
	As shown in Figure~\ref{fig:framework}, uLEAD-TabPFN consists of three modules: Representative Context Set (RCS) construction, dependency-aligned latent modeling, and uncertainty-aware scoring.
	The RCS construction module builds a compact reference set by sampling from normal training data, which serves as a in-context conditioning set for TabPFN during conditional dependency estimation.
	
	The dependency-aligned latent modeling module projects the original feature space into a latent space that is explicitly optimized for conditional dependency estimation. 
	An encoder maps input features to latent representations, and jointly performs dimensionality reduction for high-dimensional data to improve computational efficiency. 
	Conditional dependencies are then evaluated in the latent space by modeling each latent dimension using a frozen TabPFN, which conditions on the remaining latent dimensions and the RCS.
		
	In the uncertainty-aware scoring module, for each latent dimension, a lightweight multilayer perceptron (MLP), referred to as a variance network, is introduced to model input-dependent variability in dependency deviations.
	The variance network predicts a conditional scale for the target dimension given the remaining latent dimensions as input, which is used to normalize dependency deviations during anomaly scoring.
	uLEAD-TabPFN adopts a two-stage training procedure, where the dependency model is trained first, followed by the training of the variance networks using the resulting dependency deviations.
		
	These modules reflect the key design choices of uLEAD-TabPFN:
	\begin{enumerate}
		\item learning a latent representation that is explicitly aligned with conditional dependency evaluation, while also enabling dimensionality reduction for high-dimensional data;
		\item leveraging frozen TabPFN predictors for in-context conditional dependency estimation, exploiting their ability to model complex conditional relationships from limited reference contexts without task-specific retraining; and
		\item modeling conditional variability to normalize dependency deviations across regions with different intrinsic uncertainty when constructing anomaly scores.
	\end{enumerate}

	\subsection{Representative Context Set Construction}	
	The goal of this module is to construct a compact in-context reference set for conditional dependency estimation with TabPFN, thereby reducing the computational cost of in-context inference during both training and inference.
	This is achieved by replacing the full training set with a smaller representative subset that serves as the conditioning context.
	
	Given the training set of normal instances $\mathcal{X}_{\mathrm{train}}$, we construct a RCS, denoted as $\mathcal{C} \subseteq \mathcal{X}_{\mathrm{train}}$.
	The RCS is obtained via a representative sampling strategy that reduces the size of the conditioning set while preserving coverage of the underlying data distribution.
	The size of the RCS is controlled by a user-defined budget $c$.
	If $|\mathcal{X}_{\mathrm{train}}| \leq c$, the entire training set is used as the context.
	When $|\mathcal{X}_{\mathrm{train}}| > c$, the RCS is constructed using a clustering-based prototype selection strategy~\cite{feldman2011unified,bachem2018scalable}.
	Specifically, K-means clustering with a fixed number of clusters ($k=100$) is applied to partition the training data.
	The total context budget $c$ is uniformly allocated across clusters.
	Within each cluster, samples are selected to represent both central regions (points close to the cluster centroid) and more dispersed regions (points farther from the centroid), with approximately half of the samples drawn from each, subject to the overall budget constraint.
		
	By using a compact context set that is constructed once from the normal training data and kept unchanged thereafter, the RCS reduces the computational cost of in-context dependency estimation with TabPFN during both training and inference.
		
	\subsection{Dependency-Aligned Latent Modeling}	
	
	Directly estimating conditional relationships in the original feature space can be unstable and computationally inefficient, particularly in high-dimensional settings. 
	To address this, we introduce a dependency-aligned latent modeling module that learns a representation space tailored for dimension-wise conditional inference with frozen TabPFN predictors.
	
	This module consists of two components: a lightweight linear encoder and frozen TabPFN predictors.
	The encoder serves two purposes: (1) it maps observed features into latent representations that are explicitly trained to align with conditional dependency estimation, and (2) for high-dimensional data, it reduces the dimensionality of the feature space to improve computational efficiency.
	
	Frozen TabPFN predictors are employed as conditional models for dependency estimation for three reasons:
	(1) they leverage prior knowledge learned from large and diverse tabular datasets to support reliable conditional dependency estimation;
	(2) they enable dependency modeling via in-context inference without additional training, reducing computational cost; and
	(3) they avoid dataset-specific fine-tuning, simplifying the training procedure and supporting reliable dependency estimation across datasets.
	
	Conditional dependency estimation is performed in the latent space using these frozen TabPFN predictors.
	Each TabPFN predictor estimates the conditional mean of a latent dimension given the remaining latent dimensions, using the RCS as an in-context reference set.
	
	Specifically, given an input sample $\mathbf{x}_i \in \mathbb{R}^d$, a latent representation $\mathbf{z}_i \in \mathbb{R}^p$ is obtained via a linear encoder 
	\begin{equation}
		\mathbf{z}_i = W_{\mathrm{enc}} \mathbf{x}_i + b_{\mathrm{enc}}, 
	\end{equation}
	where $W_{\mathrm{enc}} \in \mathbb{R}^{p \times d}$ and $b_{\mathrm{enc}} \in \mathbb{R}^p$ are learnable parameters and $p \leq d$. 
	The encoder provides a compact representation that is trained to align with the dependency structure of normal data through the dependency-based objective. 
	Expressive nonlinear conditional relationships are captured by the frozen TabPFN model, allowing the encoder to focus on representation alignment and noise reduction rather than performing direct predictive modeling.
	
	To align the latent space with the dependency structure of normal data, uLEAD-TabPFN trains the encoder using a dependency-based objective. 
	For each latent dimension $j = 1,\dots,p$ and each training sample $i$, TabPFN is used to estimate the sample-specific conditional mean
	$ \hat{\mu}_{ij} = \mathbb{E}\!\left[ Z_j \mid Z_{-j} = \mathbf{z}_{i,-j}, Z_\mathcal{C} \right], $
	where the expectation is taken with respect to the predictive distribution induced by TabPFN with the RCS as context set. 
	This procedure is applied independently for each latent dimension.
	The dependency loss is defined as
	\begin{equation}
		\mathcal{L}_{\mathrm{dep}} = \frac{1}{N p} \sum_{i=1}^{N} \sum_{j=1}^{p} \left( z_{ij} - \hat{\mu}_{ij} \right)^2.
	\end{equation}	
	Minimizing this loss encourages normal samples to exhibit strong and consistent conditional dependencies in the latent space.
	
	To prevent degenerate solutions and preserve information from the original feature space, we include a reconstruction regularization term based on a linear decoder. Concretely, in addition to the encoder, we use a separate linear decoder
	\begin{equation}
		\hat{\mathbf{x}}_i = W_{\mathrm{dec}} \mathbf{z}_i + b_{\mathrm{dec}},
	\end{equation}
	where $W_{\mathrm{dec}} \in \mathbb{R}^{d \times p}$ and $b_{\mathrm{dec}} \in \mathbb{R}^d$ are learned independently (i.e., we do not tie decoder weights to $W_{\mathrm{enc}}^\top$).
	
	The reconstruction loss is defined using the Huber penalty, which is less sensitive to large residuals and supports stable representation learning:	
	\begin{equation}
		\mathcal{L}_{\mathrm{rec}} = \frac{1}{N}\sum_{i=1}^{N}\sum_{k=1}^{d} \rho_{\delta}\!\left(\hat{x}_{ik}-x_{ik}\right),
	\end{equation}
	where $\rho_{\delta}(u)=\frac{1}{2}u^2$ if $|u|\le\delta$ and $\rho_{\delta}(u)=\delta\left(|u|-\frac{1}{2}\delta\right)$ otherwise. We use $\delta=1$ in our implementation.	
	
	The final representation learning objective is given by
	\begin{equation}
		\mathcal{L}_{\mathrm{rl}} =
		\lambda_{\mathrm{dep}} \mathcal{L}_{\mathrm{dep}} +
		\lambda_{\mathrm{rec}} \mathcal{L}_{\mathrm{rec}},
	\end{equation}
	where $\lambda_{\mathrm{dep}}$ and $\lambda_{\mathrm{rec}}$ control the trade-off between dependency alignment and information preservation.
	
	The latent mapping uses a simple linear architecture with a single-layer linear encoder ($d \rightarrow p$) and a single linear-layer decoder ($p \rightarrow d$). 
	This minimal design avoids unnecessary modeling capacity and enables efficient training even with limited data.
		
	Overall, this module enables efficient dependency evaluation by performing dimension-wise conditional inference in a compact latent space using frozen TabPFN predictors with RCS-based in-context conditioning.
				
	\subsection{Uncertainty-Aware Scoring}
	
\begin{figure}[htbp]
		\centering
		\includegraphics[width=1\linewidth]{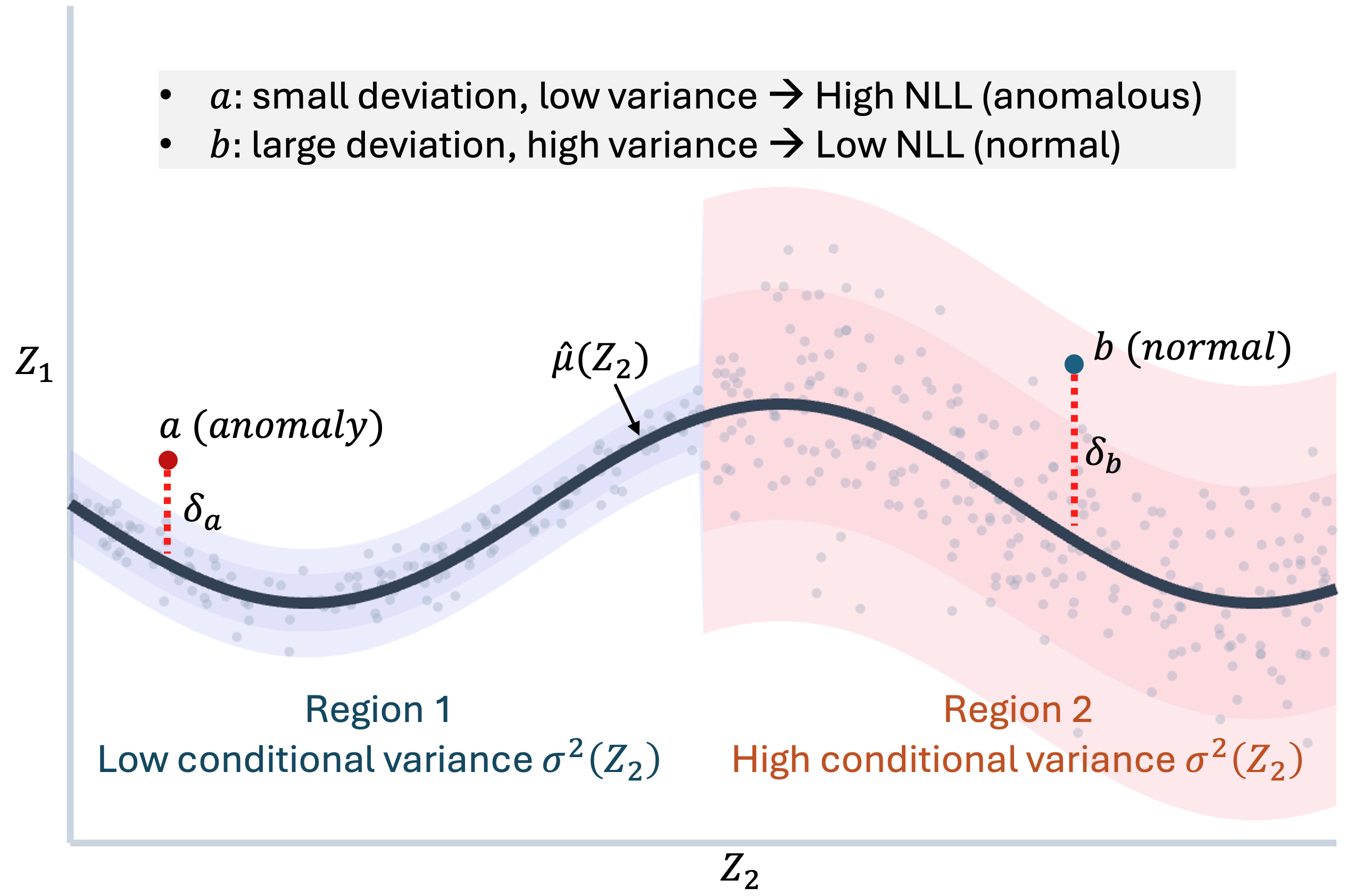}
		\caption{An illustrative example of uncertainty-aware scoring under heteroscedastic dependencies. The black curve represents the conditional mean $\hat{\mu}(Z_2)$ of latent variable $Z_1$. Region~1 exhibits low conditional variance, while Region~2 exhibits high conditional variance. Point $a$ shows a small deviation from the mean but lies in a low-variance region, resulting in a high negative log-likelihood (anomalous). Point $b$ exhibits a larger deviation but lies in a high-variance region, yielding a low negative log-likelihood (normal).}
		\label{fig:heteroscedastic_example}
	\end{figure}

	In real-world tabular data, dependency deviations alone are insufficient for reliable anomaly detection, as normal data often exhibits input-dependent variability (heteroscedasticity)~\cite{greene2012econometric}, as illustrated in Figure~\ref{fig:heteroscedastic_example}.
	Deviations of similar magnitude may therefore be benign in regions of high conditional variance but indicative of anomalies in more stable regions.
	To address this, uLEAD-TabPFN incorporates uncertainty-aware scoring by explicitly modeling the conditional variance of latent dimensions.
	
	For each latent dimension $j$, in addition to the conditional mean $\hat{\mu}_{ij}$ estimated by TabPFN, we learn an input-dependent conditional variance $\hat{\sigma}_{ij}^2$ using a lightweight variance network.
	Rather than assuming a global variance, this network models heteroscedastic residual uncertainty by predicting the conditional variance of the dependency residual
	$r_{ij} = z_{ij} - \hat{\mu}_{ij}$ as a function of the remaining latent dimensions $\mathbf{z}_{i,-j}$.
	The variance network is trained using dependency residuals computed from normal training samples encoded by the frozen encoder.
	
	Training is performed by minimizing a Gaussian negative log-likelihood objective:
	\begin{equation}
		\mathcal{L}_{\mathrm{var}} =
		\frac{1}{N p} \sum_{i=1}^{N} \sum_{j=1}^{p}
		\left[
		\frac{(z_{ij} - \hat{\mu}_{ij})^2}{\hat{\sigma}_{ij}^2}
		+ \log \hat{\sigma}_{ij}^2
		\right].
	\end{equation}
	This objective penalizes large prediction errors while discouraging trivial solutions with inflated variance, encouraging calibrated uncertainty estimates.
	
	At inference time, the anomaly score for a test sample $\mathbf{x}$ is computed by aggregating normalized dependency deviations across latent dimensions:
	\begin{equation}
		s(\mathbf{x}) =
		\frac{1}{p} \sum_{j=1}^{p}
		\frac{(z_j - \hat{\mu}_j)^2}{\hat{\sigma}_j^2}.
	\end{equation}
	
	The conditional variance $\hat{\sigma}_{ij}^2$ is predicted using a lightweight variance network with a fixed and deliberately simple architecture. 
	For each target latent dimension, the network takes the remaining $p-1$ latent dimensions $\mathbf{z}_{i,-j}$ as input.
	The network is implemented as a MLP with two hidden layers, consisting of a fully connected layer with $64$ neurons followed by a ReLU activation, a second fully connected layer with $32$ neurons followed by a ReLU activation, and a final linear layer with a single output neuron that predicts $\log \hat{\sigma}_{ij}^2$.
	Dropout with a rate of $0.1$ is applied after each hidden layer to improve robustness.
	This architecture introduces minimal additional modeling capacity while remaining sufficient to capture input-dependent residual uncertainty.
	
	By normalizing dependency deviations using conditional variance estimated from normal data, uLEAD-TabPFN evaluates test samples relative to the expected variability of normal conditional structure, improving robustness and reducing false positives under heteroscedastic and multi-modal normal behavior.
		
	Figure~\ref{fig:heteroscedastic_example} illustrates the motivation for uncertainty-aware scoring under heteroscedastic dependencies. The black curve represents the conditional mean, with Region~1 exhibiting low conditional variance and Region~2 exhibiting high conditional variance. Although point $a$ has a smaller deviation from the mean than point $b$, it lies in a low-variance region and therefore yields a higher negative log-likelihood. In contrast, point $b$ lies in a high-variance region where larger deviations are expected, resulting in a lower negative log-likelihood. This example highlights that raw dependency deviations are insufficient under heteroscedastic noise, whereas variance-normalized deviations enable consistent anomaly scoring.
	
	\subsection{Computational Complexity}
	\label{sec:complexity}
	We analyze the computational complexity of uLEAD-TabPFN with respect to input dimensionality $d$, latent dimensionality $p$, and the size of the representative context set $|\mathcal{C}|$.
	
	\paragraph{Training Complexity.} 
	Training involves learning the encoder and the variance networks, while the TabPFN predictors remain frozen. For each training sample, dependency-aligned representation learning requires conditional estimation over $p$ latent dimensions using the fixed RCS. As a result, the overall training complexity scales linearly with the number of training samples $N$, the latent dimensionality $p$, and the context size $|\mathcal{C}|$. In practice, both $p$ and $|\mathcal{C}|$ are fixed and substantially smaller than $d$ and $N$, making training efficient even for high-dimensional tabular data.
	
	\paragraph{Inference Complexity.}
	At inference time, each test sample is processed independently. The dominant cost arises from in-context dependency estimation using TabPFN, which scales linearly with $p$ and $|\mathcal{C}|$. No dataset-specific fine-tuning and iterative optimization is required at test time, enabling efficient and stable deployment.
	
	\paragraph{Memory and Practical Considerations.}
	The memory footprint of uLEAD-TabPFN is determined primarily by the storage of the context set $\mathcal{C}$ and the parameters of the encoder and variance networks. Since the context set is fixed after construction, memory usage remains constant during inference. In practice, the context size $|\mathcal{C}|$ provides a controllable trade-off between computational efficiency and representational coverage and is treated as a user-defined parameter.
	
	Overall, by combining a compact RCS, lightweight trainable components, and a frozen TabPFN, uLEAD-TabPFN achieves favorable scalability and practical efficiency for dependency-based anomaly detection in high-dimensional tabular settings.
	
	\begin{table*}[htbp]
		\centering
		\caption{
			Summary of ROC-AUC and PR-AUC over 57 ADBench datasets, reported for the top-10 performing methods.
			Datasets are grouped by feature dimensionality: Low ($d \leq 20$, 24 datasets), Medium ($20 < d \leq 100$, 18 datasets), High ($d > 100$, 15 datasets), and All.
			Results are shown as mean performance with rank in parentheses.
			For uLEAD-TabPFN, relative changes compared to the best-performing baseline (\%B) and the average performance across all baselines (\%A) are additionally reported.
		}
		\label{tab:main_results}
		\resizebox{1\textwidth}{!}{%
			\begin{tabular}{c|l|c|ccccccccc}
				\toprule
				\rowcolor{darkgray}
				\textbf{Metric} & \textbf{Dimensionality} & uLEAD-TabPFN & FoMo-0D & DTE-NP & kNN & ICL & LOF & CBLOF & FeatBag & SLAD & DDPM \\
				\midrule
				\multirow{4}{*}{\textbf{ROC-AUC}}
				& High &
				{73.98 (1)  [+2.84\%B, +19.56\%A]} &
				65.46 (13) & 71.94 (2) & 70.55 (6) & 71.14 (3) & 71.05 (4) & 67.49 (9) & 70.85 (5) & 66.36 (11) & 66.98 (10) \\
				
				& Medium &
				{80.13 (1)  [+1.47\%B, +15.59\%A]} &
				78.24 (4) & 78.97 (2) & 77.94 (5) & 78.64 (3) & 74.59 (10) & 76.24 (7) & 74.50 (11) & 76.93 (6) & 73.95 (13) \\
				
				& Low &
				{85.51 (7)  [-5.14\%B, +10.49\%A]} &
				90.14 (1) & 89.33 (3) & 89.48 (2) & 85.03 (8) & 86.15 (5) & 86.12 (6) & 82.63 (12) & 84.64 (9) & 87.73 (4) \\
				
				& All &
				{80.78 (3)  [-0.87\%B, +14.16\%A]} &
				79.89 (4) & 81.49 (1) & 80.86 (2) & 79.35 (5) & 78.53 (6) & 78.10 (7) & 76.96 (10) & 77.40 (9) & 77.92 (8) \\
				
				\midrule\midrule
				
				\rowcolor{darkgray}
				\textbf{Metric} & \textbf{Dimensionality} & uLEAD-TabPFN & DDAE & FoMo-0D & DTE-NP & kNN & ICL & LOF & SLAD & DDPM & OCSVM \\
				\midrule
				\multirow{4}{*}{\textbf{PR-AUC}}
				& High &
				{40.24 (2)  [-1.26\%B, +45.50\%A]} &
				40.76 (1) & 30.55 (11) & 34.45 (6) & 33.54 (7) & 37.59 (3) & 35.15 (4) & 31.24 (10) & 29.82 (13) & 29.50 (14) \\
				
				& Medium &
				{57.37 (3)  [-8.93\%B, +20.43\%A]} &
				62.99 (1) & 55.46 (5) & 57.04 (4) & 54.82 (7) & 57.49 (2) & 54.32 (8) & 55.20 (6) & 52.80 (10) & 51.77 (11) \\
				
				& Low &
				{62.67 (7)  [-15.82\%B, +17.04\%A]} &
				73.02 (2) & 74.45 (1) & 71.37 (4) & 71.42 (3) & 65.63 (6) & 60.95 (11) & 61.79 (9) & 66.24 (5) & 62.47 (8) \\
				
				& All &
				{55.09 (6)  [-10.21\%B, +22.79\%A]} &
				61.36 (1) & 56.90 (3) & 57.13 (2) & 56.21 (4) & 55.68 (5) & 52.07 (8) & 51.67 (9) & 52.41 (7) & 50.41 (10) \\
				
				\bottomrule
			\end{tabular}
		}
	\end{table*}
	
	\section{Experiments}
	\label{sec:experiments}
	
	\subsection{Experimental Setup}
	\label{sec:exp_setup}
	We evaluate uLEAD-TabPFN on ADBench~\cite{han2022adbench}, a comprehensive benchmark for tabular anomaly detection, following its standardized semi-supervised evaluation protocol as adopted in prior work~\cite{han2022adbench,livernochediffusion,fomo0d2024zeroshoot,sattarov2025diffusion}.
	ADBench consists of 57 tabular anomaly detection datasets spanning a wide range of dimensionalities, sample sizes, and anomaly ratios, all of which are used in our experiments.
	All features are normalized using the RCS mean and standard deviation.
	Dataset statistics are summarized in Table~\ref{tab:datasets} in Appendix~\ref{app:datasets}.
	
	We compare uLEAD-TabPFN against 26 baseline methods, including all 14 semi-supervised baselines provided by ADBench~\cite{han2022adbench}, an additional 10 baselines reported in~\cite{livernochediffusion}, and two recent methods not included in these benchmarks: a PFN-based anomaly detection approach~\cite{fomo0d2024zeroshoot} and a diffusion-based method~\cite{sattarov2025diffusion}.
	Together, these baselines cover a broad range of anomaly detection paradigms, including classical proximity-based methods, dependency-based approaches, deep learning models, and recent foundation- and diffusion-based techniques.
	We report ROC-AUC and PR-AUC, averaged over five runs with different random seeds.
	Detailed baseline descriptions and implementation details are provided in Appendix~\ref{app:baselines}.
	
	uLEAD-TabPFN is designed to minimize hyperparameter tuning. The TabPFN used uLEAD-TabPFN is kept frozen, and only lightweight components, including a linear encoder and variance networks, are trained. The primary design choice is the latent dimensionality $p$: for datasets with $d \leq 100$, $p$ is set to $d$, while for $d > 100$, it is fixed to $100$. All other parameters are fixed across datasets and seeds.
	
	The encoder is trained for 20 epochs using Adam with no weight decay and a batch size of 1024, with the learning rate scheduled from $5\times10^{-4}$ to $10^{-6}$. 
	For uncertainty-aware scoring, one variance network is trained per latent dimension for 50 epochs using Adam with a learning rate of $10^{-3}$ and weight decay of $10^{-4}$. 
	The RCS budget is set to 500. 
	The weights of the dependency loss $\lambda_{\mathrm{dep}}$ and the reconstruction loss $\lambda_{\mathrm{rec}}$ are set to 1 and 0.3, respectively.
				
	\subsection{Main Results}
	\label{sec:main_results}
	
	Table~\ref{tab:main_results} summarizes the ROC-AUC and PR-AUC performance of uLEAD-TabPFN and top-performing baseline methods across 57 datasets, grouped by feature dimensionality.
	For the baseline methods, most results are taken from~\cite{livernochediffusion}, while results for FoMo-0D and DDAE are obtained from their respective original publications.
	Results are reported as mean performance with the corresponding rank among all baseline methods. 
	For uLEAD-TabPFN, we additionally report relative performance changes with respect to the best-performing baseline (denoted as \%B) and the average performance across all baselines (denoted as \%A).	
	Note that comparisons involving DDAE are reported on PR-AUC only, as the original paper does not include the ROC-AUC of each dataset. Full per-dataset results are provided in Appendix~\ref{app:results}.
	
	Across both evaluation metrics, uLEAD-TabPFN demonstrates strong performance relative to the SOTA baselines, particularly on medium- and high-dimensional datasets.
	Under ROC-AUC, uLEAD-TabPFN achieves the best average rank in both dimensionality categories, with improvements of approximately $+1.5\%$ and $+2.8\%$ over the best-performing baseline, and gains of about $+16\%$ and $+20\%$ over the average baseline performance, respectively.
	A similar pattern is observed for PR-AUC. In the medium-dimensional setting, uLEAD-TabPFN improves over the average baseline by around $20\%$ (rank 3), while in the high-dimensional regime the improvement exceeds $45\%$ (rank 2). 
	In low-dimensional datasets, uLEAD-TabPFN performs comparably to strong baselines, indicating competitive performance in regimes where simpler methods are already effective.
	
	We further assess the statistical significance of the results using Wilcoxon signed-rank tests between uLEAD-TabPFN and each baseline within each dimensionality group. 
	As shown in Table~\ref{tab:wilcoxon_pvalues} in Appendix~\ref{app:results}, among all the 57 datasets, uLEAD-TabPFN achieves 17 out of 25 significant wins in ROC-AUC and 16 out of 26 in PR-AUC. 
	For high-dimensional datasets ($d>100$), uLEAD-TabPFN attains 20 significant wins in both ROC-AUC and PR-AUC. 
	Fewer statistically significant differences are observed in low-dimensional settings, with 6/25 wins for ROC-AUC and 9/26 for PR-AUC.
	
	Overall, these results demonstrate that uLEAD-TabPFN achieves consistently strong performance across diverse tabular datasets, with clear advantages emerging as feature dimensionality and dependency complexity increase.
	
	\subsection{Performance Complementarity Analysis}
	\label{sec:uniqueness}
	
	We observed that uLEAD-TabPFN exhibits particularly strong performance on datasets where many existing methods struggle.
	For example, on the InternetAds dataset, baseline methods achieve ROC-AUC scores ranging from approximately $53\%$ to $76\%$, whereas uLEAD-TabPFN attains about $90\%$.
	Similarly, on the Wilt dataset, baseline methods reach ROC-AUC values between $32\%$ and $75\%$, while uLEAD-TabPFN achieves $97\%$.
	These results indicate that uLEAD-TabPFN captures anomaly characteristics that are not well addressed by conventional approaches.	
	\begin{figure}[htbp]
		\centering
		\includegraphics[width=\columnwidth]{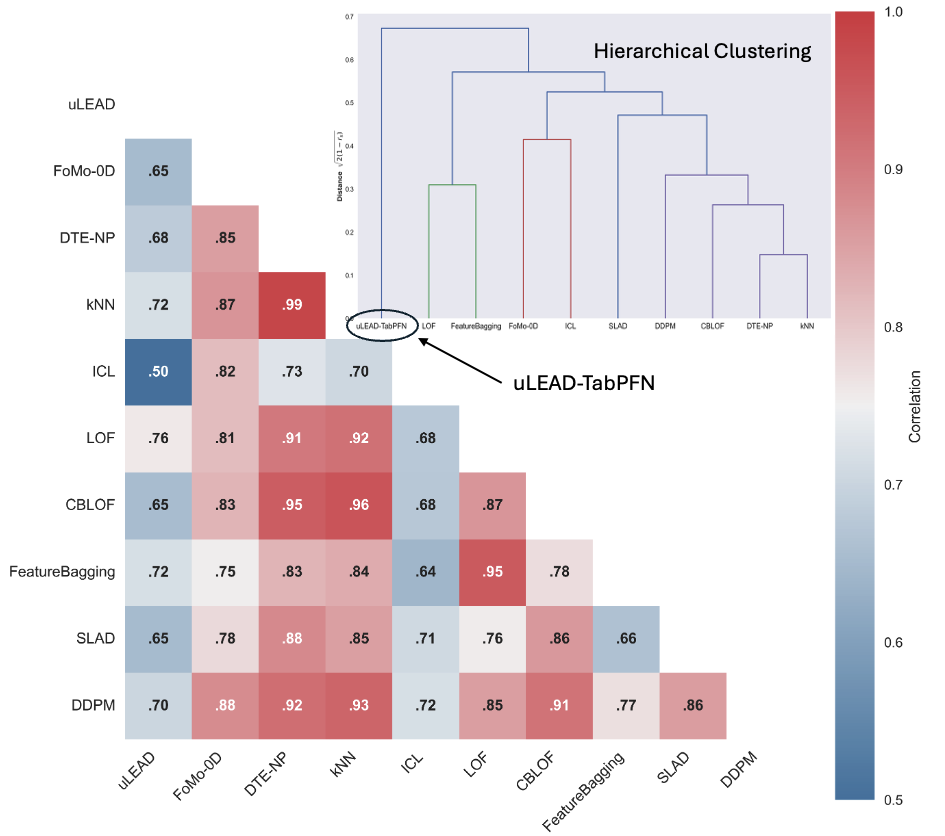}
		\caption{
			Performance complementarity analysis of top-performing anomaly detection methods on 57 ADBench datasets.
			The heatmap shows pairwise Spearman rank correlations of ROC-AUC scores, with hierarchical clustering applied to reveal similarity patterns among methods.
			uLEAD-TabPFN exhibits consistently low correlation with other approaches and forms a distinct cluster, indicating complementary behavior.
		}		
		\label{fig:complementary}
	\end{figure}
	
	Motivated by this observation, we analyze dataset-level performance patterns of uLEAD-TabPFN relative to top-performing baselines using correlation and clustering analyses based on ROC-AUC scores.
	As shown in Figure~\ref{fig:complementary}, uLEAD-TabPFN exhibits consistently lower Spearman rank correlation with most baseline methods than the correlations observed among the baselines themselves.
	Specifically, its average correlation with other methods is $0.67$, compared to values above $0.83$ for several strong baselines such as DTE-NP, kNN, DDPM, and LOF.
	In contrast, many baseline methods form tightly correlated groups, indicating shared inductive biases and sensitivity to similar dataset characteristics.
		
	Hierarchical clustering based on a metric-consistent distance derived from rank correlations~\cite{mantegna1999hierarchical} reveals a consistent structural pattern, with uLEAD-TabPFN separating from the dominant baseline clusters rather than merging into any of them.
	Together, these observations indicate that uLEAD-TabPFN captures anomaly evidence that is largely non-overlapping with that of the baseline methods, highlighting its complementary behavior.
			
	\subsection{Ablation Study}
	\label{sec:ablation}
	
	We conduct ablation studies to assess the contribution of key components in uLEAD-TabPFN:
	\begin{itemize}
		\item the dependency-aligned linear encoder, evaluated via the ablation variant \textit{uLEAD-Original};
		\item the dependency prediction model based on TabPFN, evaluated via the ablation variant \textit{uLEAD-CART}; and
		\item uncertainty-aware scoring, evaluated via the ablation variant \textit{LEAD-TabPFN}.
	\end{itemize}
	
	For computational reasons, the \textit{uLEAD-Original} variant is evaluated only on datasets with $d \leq 100$, as removing the linear encoder requires performing dependency prediction directly in the original feature space, which becomes computationally prohibitive for higher-dimensional datasets.
	We use CART as the alternative dependency predictor because it achieved strong and stable performance among classical models, as reported in~\cite{lu2025dependency}.
	\begin{table}[htbp]
		\centering
		\caption{
			Ablation study on ROC-AUC and PR-AUC over the ADBench datasets (mean $\pm$ std).
			\textit{LEAD-TabPFN}: without uncertainty-aware scoring;
			\textit{uLEAD-Original}: without the linear encoder;
			\textit{uLEAD-CART}: using CART instead of TabPFN for dependency prediction.
			uLEAD-Original is evaluated only on datasets with $d \leq 100$ due to computational constraints.
		}		
		\label{tab:ablation}
		\resizebox{0.9\columnwidth}{!}{%
			\begin{tabular}{llcc}
				\toprule
				\rowcolor{darkgray}
				\textbf{Metric} & \textbf{Method} & \textbf{$d \leq 100$} & \textbf{All} \\
				\midrule
				
				\multirow{4}{*}{\textbf{ROC-AUC}}
				& uLEAD-TabPFN    & \textbf{83.21$\pm$13.49} & \textbf{80.78$\pm$14.39} \\
				& LEAD-TabPFN     & 81.33$\pm$15.15         & 79.20$\pm$15.56 \\
				& uLEAD-CART      & 78.49$\pm$15.63         & 76.90$\pm$15.49 \\
				& uLEAD-Original  & 71.98$\pm$19.54         & -- \\
				\midrule
				
				\multirow{4}{*}{\textbf{PR-AUC}}
				& uLEAD-TabPFN    & \textbf{60.40$\pm$24.70} & \textbf{55.09$\pm$28.55} \\
				& LEAD-TabPFN     & 52.87$\pm$28.91          & 49.47$\pm$30.43 \\
				& uLEAD-CART      & 46.93$\pm$29.23          & 44.44$\pm$29.94 \\
				& uLEAD-Original  & 43.55$\pm$28.71          & -- \\
				\bottomrule
			\end{tabular}
		}
	\end{table}

	Table~\ref{tab:ablation} summarizes the ablation results. 
	Across both ROC-AUC and PR-AUC, uLEAD-TabPFN consistently achieves the best mean performance, while also exhibiting comparatively lower variability across datasets. Removing the dependency-aligned linear encoder (\textit{uLEAD-Original}) leads to the most pronounced performance degradation in both metrics and substantially increases variability across datasets. 
	This observation indicates that aligning the latent representation with conditional dependency modeling is critical to the effectiveness of uLEAD-TabPFN.
	
	Replacing TabPFN with a CART predictor (\textit{uLEAD-CART}) also results in clear performance degradation and relatively high variability, highlighting the importance of utilizing TabPFNs for dependency modeling. In contrast, removing uncertainty-aware scoring (\textit{LEAD-TabPFN}) leads to a more moderate but consistent decrease in mean performance, accompanied by increased variability, particularly for PR-AUC. 
	This suggests that uncertainty-aware scoring primarily contributes to robustness and stability across heterogeneous datasets, rather than being the primary source of performance gains.
	
	Overall, the ablation results indicate that uLEAD-TabPFN benefits most from dependency-aligned representation learning and the use of TabPFNs for dependency prediction, while uncertainty-aware scoring provides complementary stabilization effects.

	\subsection{Sensitivity Study}
	\label{sec:latent_dim}	
	\begin{figure}[htbp]
		\centering
		\includegraphics[width=1\columnwidth]{sensitivity.png}
		\caption{Sensitivity of uLEAD-TabPFN to latent dimensionality for datasets with $d \leq 100$. Average ROC-AUC (blue curve and left axis) and PR-AUC (red curve and right axis) are shown as functions of the latent dimension cap $p$. Performance improves rapidly for small $p$ and stabilizes once a moderate latent dimensionality is reached.}
		\label{fig:latent_dim}
	\end{figure}
	
	uLEAD-TabPFN is deliberately designed with a minimal set of hyperparameters, most of which are fixed across all experiments (as described in Section~\ref{sec:exp_setup}) to avoid dataset-specific tuning.
	As a result, we focus our sensitivity analysis on the primary structural hyperparameter, namely the latent dimensionality $p$.
	We restrict this analysis to datasets with $d \leq 100$ to avoid confounding the effect of $p$ with the aggressive dimensionality reduction that would be required for datasets with $d > 100$.
	
	Figure~\ref{fig:latent_dim} reports the average ROC-AUC and PR-AUC over datasets with $d \leq 100$ as $p$ varies.
	Performance improves rapidly as $p$ increases from small values and stabilizes around $p \approx 40$, with only marginal gains observed for larger values up to $p = 100$.
	These results indicate that uLEAD-TabPFN is not sensitive to the exact choice of latent dimensionality once a moderate number of latent variables is retained.
		
	Overall, these results show that uLEAD-TabPFN does not require fine-grained tuning of latent dimensionality and maintains stable performance across a broad range of values, supporting its practicality for large-scale and heterogeneous tabular anomaly detection.
	
	\subsection{Case Study on the Wilt Dataset}
	\label{sec:case_study}
	We present a case study on the \textit{Wilt} dataset~\cite{wilt_285} to illustrate how uLEAD-TabPFN operates in practice, from representation learning to dependency-based scoring and interpretation. 
	Wilt is a widely used tabular anomaly detection benchmark derived from remote sensing imagery, where diseased trees correspond to anomalous instances. 
	The dataset contains five continuous spectral and texture-related features with clear physical semantics and 4,819 samples (4,562 normal and 257 anomalous), making it well suited for interpretability analysis.
	
	On the Wilt dataset, uLEAD-TabPFN achieves a ROC-AUC of 97.5\% and a PR-AUC of 90.8\%.
	In contrast, most baseline methods perform poorly, with ROC-AUCs largely below 75\% and PR-AUCs below 20\%.
	Ablation results (Section~\ref{sec:ablation}) show that removing the linear encoder (LEAD-Original, 47\% ROC-AUC) or replacing TabPFN with a CART predictor (uLEAD-CART, 64\% ROC-AUC) leads to substantial performance degradation, whereas removing uncertainty-aware scoring has a smaller effect (LEAD-TabPFN, 87\% ROC-AUC).
	This suggests that the dominant gains arise from dependency-aligned representation learning and TabPFN-based dependency modeling.	
	
	\begin{figure}[htbp]
		\centering
		\includegraphics[width=\columnwidth]{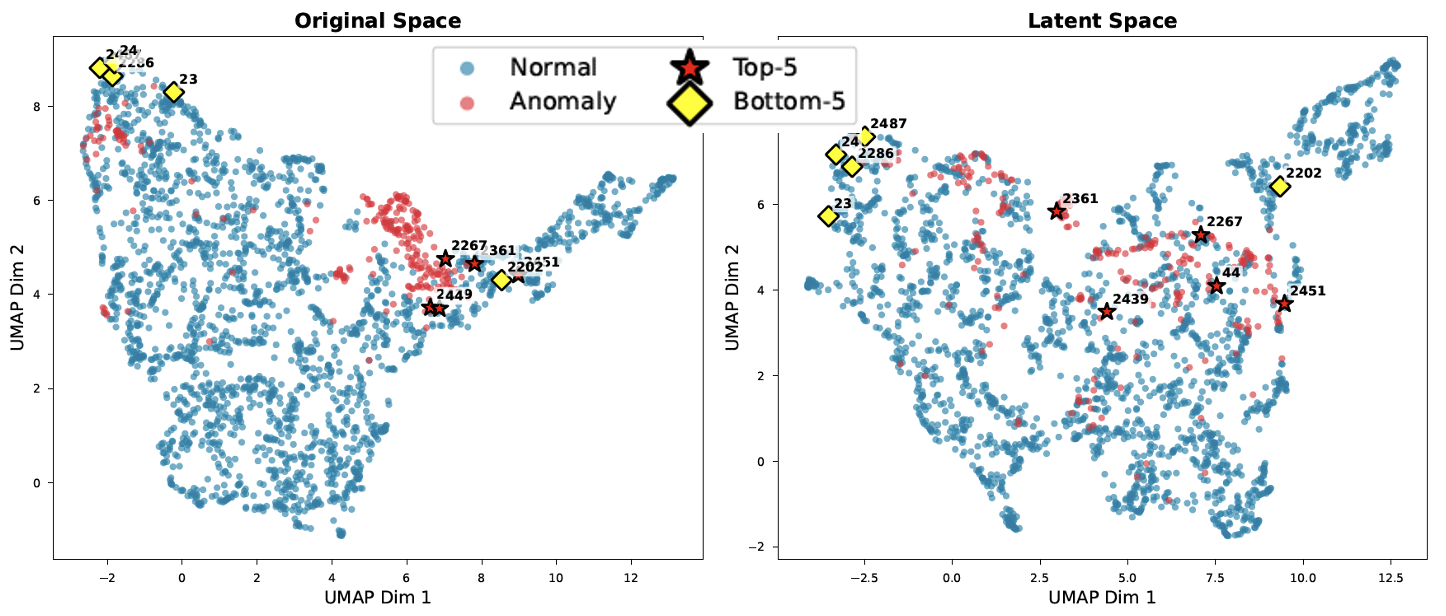}
		\caption{
			UMAP visualization of the Wilt dataset in the original feature space (left) and the latent space (right).
			While normal and anomalous samples remain overlapping in both spaces, the latent space reveals more structured organization of anomalous samples along the data manifold, consistent with the learned dependency-aware representation.
		}
		\label{fig:wilt_umap}
	\end{figure}
	
	\paragraph{Latent Space Analysis.}
	Figure~\ref{fig:wilt_umap} shows UMAP visualizations of the Wilt dataset in the original and latent feature spaces, comparing the data organization before and after latent transformation.
	In both representations, normal and anomalous samples exhibit substantial overlap, indicating the absence of clear global geometric separation.
	To further understand how the latent transformation affects the data structure, we analyze dependency-related and neighborhood-level characteristics in both spaces.
	
	Quantitatively, the mean absolute Pearson correlation between feature pairs increases from $0.239$ in the original space to $0.531$ in the latent space, suggesting that dependency-related structure is more explicitly captured after latent transformation.
	Consistently, the average conditional $R^2$, computed using a frozen TabPFN as a fixed conditional estimator, increases from $0.649$ to $0.991$, indicating that conditional relationships become substantially more predictable and stable in the latent space.
	
	Importantly, despite these changes, the lack of global separation implies that distance- or density-based anomaly detection methods remain ineffective, as they primarily rely on marginal proximity or density cues that are weak in both the original and latent spaces.
	In contrast, the strengthened conditional dependency structure in the latent space enables uLEAD-TabPFN to detect anomalies through violations of learned dependencies rather than global geometric separation.
		
	\paragraph{Anomaly Interpretation.}
	uLEAD-TabPFN provides interpretability through its scoring mechanism. Anomaly scores are decomposed into latent-dimension contributions that correspond to violated conditional dependencies. Because the latent representation is obtained via a linear encoder, these latent-level deviations can be directly mapped back to the original input features, enabling faithful feature-level attribution without post-hoc explanation.
	
	For high-scoring anomalies in the Wilt dataset, the largest contributions typically arise from panchromatic intensity variation (SD-Pan) and near-infrared reflectance (Mean-NIR). 
	The texture feature GLCM-Pan exhibits a moderate contribution, while the visible-band features Mean-R and Mean-G contribute less to the overall anomaly score.
	This feature-level attribution pattern is consistently observed across other high-scoring anomalies.
	
	A complete case analysis, including per-dimension statistics and additional examples, is provided in Appendix~\ref{app:case}.
	
	\section{Conclusion}
	\label{conclusion}
	
	We have proposed uLEAD-TabPFN, a dependency-based anomaly detection framework for tabular data that identifies anomalies as violations of conditional dependencies among variables.
	The framework integrates dependency-aligned latent representation learning, in-context conditional dependency estimation using pre-trained TabPFNs, and uncertainty-aware scoring.
	By learning a low-dimensional latent space explicitly aligned with conditional dependency evaluation, uLEAD-TabPFN alleviates the challenges posed by high-dimensional feature spaces.
	Moreover, by leveraging the expressive conditional modeling capability of pre-trained TabPFNs through in-context inference, the framework avoids dataset-specific dependency model training while enabling robust and scalable dependency-based anomaly detection.
	
	Experiments on 57 datasets from ADBench show that uLEAD-TabPFN is particularly effective in medium- and high-dimensional settings, while maintaining strong overall performance relative to a broad range of SOTA methods.
	Ablation studies further confirm that PFN-based conditional modeling is central to these performance gains, while uncertainty-aware scoring improves robustness under heterogeneous normal data.
	Beyond aggregate performance, correlation and clustering analyses indicate that uLEAD-TabPFN exhibits complementary behavior relative to existing methods, reflecting a distinct anomaly detection mechanism rather than incremental refinement.
		
	While uLEAD-TabPFN performs consistently well overall, variability on a small subset of datasets indicates opportunities for further improvement. Future work will explore more explicit handling of complex normal data distributions and principled integration of dependency-based anomaly evidence with other detection criteria.
	
	Overall, uLEAD-TabPFN provides an effective, efficient, and practical solution for tabular anomaly detection, particularly in challenging settings characterized by high dimensionality and complex feature dependencies, while achieving strong empirical performance with minimal tuning overhead.
		
	%\section{Acknowledgments}
	%
	% \begin{acks}
		% \end{acks}
	\bibliographystyle{ACM-Reference-Format}
	\bibliography{reference}
	\appendix
	
	\section{Appendix: Datasets Details}
	\label{app:datasets}
	The dataset information used in the experiments is presented in Table~\ref{tab:datasets}.
	
	\begin{table*}[htbp]
		\centering
		\small
		\resizebox{0.65\textwidth}{!}{
			\begin{tabular}{lrrrrl}
				\toprule
				Dataset & \#Samples & \#Features & \#Anomalies & \%Anomalies & Category \\
				\midrule
				ALOI & 49534 & 27 & 1508 & 3.04 & Image \\
				annthyroid & 7200 & 6 & 534 & 7.42 & Healthcare \\
				backdoor & 95329 & 196 & 2329 & 2.44 & Network \\
				breastw & 683 & 9 & 239 & 34.99 & Healthcare \\
				campaign & 41188 & 62 & 4640 & 11.27 & Finance \\
				cardio & 1831 & 21 & 176 & 9.61 & Healthcare \\
				Cardiotocography & 2114 & 21 & 466 & 22.04 & Healthcare \\
				celeba & 202599 & 39 & 4547 & 2.24 & Image \\
				census & 299285 & 500 & 18568 & 6.20 & Sociology \\
				cover & 286048 & 10 & 2747 & 0.96 & Botany \\
				donors & 619326 & 10 & 36710 & 5.93 & Sociology \\
				fault & 1941 & 27 & 673 & 34.67 & Physical \\
				fraud & 284807 & 29 & 492 & 0.17 & Finance \\
				glass & 214 & 7 & 9 & 4.21 & Forensic \\
				Hepatitis & 80 & 19 & 13 & 16.25 & Healthcare \\
				http & 567498 & 3 & 2211 & 0.39 & Web \\
				InternetAds & 1966 & 1555 & 368 & 18.72 & Image \\
				Ionosphere & 351 & 32 & 126 & 35.90 & Oryctognosy \\
				landsat & 6435 & 36 & 1333 & 20.71 & Astronautics \\
				letter & 1600 & 32 & 100 & 6.25 & Image \\
				Lymphography & 148 & 18 & 6 & 4.05 & Healthcare \\
				magic.gamma & 19020 & 10 & 6688 & 35.16 & Physical \\
				mammography & 11183 & 6 & 260 & 2.32 & Healthcare \\
				mnist & 7603 & 100 & 700 & 9.21 & Image \\
				musk & 3062 & 166 & 97 & 3.17 & Chemistry \\
				optdigits & 5216 & 64 & 150 & 2.88 & Image \\
				PageBlocks & 5393 & 10 & 510 & 9.46 & Document \\
				pendigits & 6870 & 16 & 156 & 2.27 & Image \\
				Pima & 768 & 8 & 268 & 34.90 & Healthcare \\
				satellite & 6435 & 36 & 2036 & 31.64 & Astronautics \\
				satimage-2 & 5803 & 36 & 71 & 1.22 & Astronautics \\
				shuttle & 49097 & 9 & 3511 & 7.15 & Astronautics \\
				skin & 245057 & 3 & 50859 & 20.75 & Image \\
				smtp & 95156 & 3 & 30 & 0.03 & Web \\
				SpamBase & 4207 & 57 & 1679 & 39.91 & Document \\
				speech & 3686 & 400 & 61 & 1.65 & Linguistics \\
				Stamps & 340 & 9 & 31 & 9.12 & Document \\
				thyroid & 3772 & 6 & 93 & 2.47 & Healthcare \\
				vertebral & 240 & 6 & 30 & 12.50 & Biology \\
				vowels & 1456 & 12 & 50 & 3.43 & Linguistics \\
				Waveform & 3443 & 21 & 100 & 2.90 & Physics \\
				WBC & 223 & 9 & 10 & 4.48 & Healthcare \\
				WDBC & 367 & 30 & 10 & 2.72 & Healthcare \\
				Wilt & 4819 & 5 & 257 & 5.33 & Botany \\
				wine & 129 & 13 & 10 & 7.75 & Chemistry \\
				WPBC & 198 & 33 & 47 & 23.74 & Healthcare \\
				yeast & 1484 & 8 & 507 & 34.16 & Biology \\
				CIFAR10 & 5263 & 512 & 263 & 5.00 & Image \\
				FashionMNIST & 6315 & 512 & 315 & 5.00 & Image \\
				MNIST-C & 10000 & 512 & 500 & 5.00 & Image \\
				MVTec-AD & 5354 & 512 & 1258 & 23.50 & Image \\
				SVHN & 5208 & 512 & 260 & 5.00 & Image \\
				Agnews & 10000 & 768 & 500 & 5.00 & NLP \\
				Amazon & 10000 & 768 & 500 & 5.00 & NLP \\
				Imdb & 10000 & 768 & 500 & 5.00 & NLP \\
				Yelp & 10000 & 768 & 500 & 5.00 & NLP \\
				20news & 11905 & 768 & 591 & 4.96 & NLP \\
				\bottomrule
			\end{tabular}
		}
		\caption{Summary of datasets used for evaluation.}
		\label{tab:datasets}
	\end{table*}
	
	\section{Appendix: Baseline Methods}
	\label{app:baselines}
	We compare uLEAD-TabPFN against a total of 26 baseline methods drawn from three sources. First, we include all baseline methods provided in the ADBench benchmark~\cite{han2022adbench}. Second, we incorporate additional baselines evaluated by Livernoche et al.~\cite{livernochediffusion}, which extend ADBench with recent deep learning-based, generative, and diffusion-based anomaly detection methods. 
	Third, we additionally include two recently proposed methods that are not part of the above benchmarks, namely a PFN-based anomaly detection method, FoMo-0D~\cite{fomo0d2024zeroshoot}, and a diffusion-based tabular anomaly detection approach, DAEE~\cite{sattarov2025diffusion}.
	
	The evaluated baselines span a broad range of anomaly detection paradigms, including conventional, deep learning-based, generative, diffusion-based, and foundation model–based approaches. Specifically, the conventional baselines include CBLOF~\cite{he2003cblof}, COPOD~\cite{li2020copod}, ECOD~\cite{li2022ecod}, FeatureBagging~\cite{lazarevic2005featurebagging}, HBOS~\cite{goldstein2012hbos}, Isolation Forest~\cite{liu2008iforest}, kNN~\cite{ramaswamy2000knn}, LODA~\cite{pevny2016loda}, LOF~\cite{breunig2000lof}, MCD~\cite{fauconnier2009mcd}, OCSVM~\cite{scholkopf1999ocsvm}, and PCA~\cite{shyu2003pca}. Deep learning–based baselines include DeepSVDD~\cite{ruff2018deepsvdd} and DAGMM~\cite{zong2018dagmm}. Following Livernoche et al.~\cite{livernochediffusion}, we further include recent deep learning-based methods such as DROCC~\cite{goyal2020drocc}, GOAD~\cite{bergman2020goad}, ICL~\cite{shenkar2022icl}, SLAD~\cite{xu2023slad}, and DIF~\cite{xu2023dif}, as well as generative and diffusion-based approaches including normalizing flows with planar flows~\cite{rezende2015planar}, variational autoencoders (VAE)~\cite{kingma2013vae}, generative adversarial networks (GAN)~\cite{goodfellow2014gan}, and diffusion-based models using DDPMs~\cite{livernochediffusion,sattarov2025diffusion}. 
	In addition, we include the PFN-based anomaly detection method FoMo-0D~\cite{fomo0d2024zeroshoot}. Overall, our evaluation comprises 26 baseline methods.
	
	For a fair and consistent comparison, the experimental results for all semi-supervised methods included in ADBench~\cite{han2022adbench} and the additional 10 baselines introduced by Livernoche et al.~\cite{livernochediffusion} are taken directly from~\cite{livernochediffusion}. The results for FoMo-0D~\cite{fomo0d2024zeroshoot} and the diffusion-based method of~\cite{sattarov2025diffusion} are reported from their respective original papers.
		
	\section{Appendix: Implementation Details} 
	\label{app:implementation}
	The algorithms for training and inference with uLEAD-TabPFN are presented in Algorithms~\ref{alg:train_ulead} and~\ref{alg:infer_ulead}, respectively.	
	
	\begin{algorithm}[t]
		\caption{Training uLEAD-TabPFN}
		\label{alg:train_ulead}
		\begin{algorithmic}[1]
			\Require Normal training set $\mathcal{X}_{\mathrm{train}}=\{\mathbf{x}_i\}_{i=1}^N$, context budget $c$, latent dimension cap $p$
			\Require Frozen TabPFN conditional predictor $\mathrm{TabPFN}(\cdot)$
			\Ensure Trained encoder $f_\theta$, trained variance networks $\{v_{\phi_j}\}_{j=1}^p$, context set $\mathcal{C}$, normalization stats $(\mu_{\mathcal{C}},\sigma_{\mathcal{C}})$
			
			\vspace{2pt}
			\State \textbf{Construct Representative Context Set:}
			\State $\mathcal{C} \gets \mathrm{BuildContext}(\mathcal{X}_{\mathrm{train}}, c)$ \Comment{e.g., KMeans-based selection}
			\State Compute $(\mu_{\mathcal{C}},\sigma_{\mathcal{C}})$ on $\mathcal{C}$
			\State Normalize $\tilde{\mathbf{x}}\gets (\mathbf{x}-\mu_{\mathcal{C}})\oslash\sigma_{\mathcal{C}}$ for all $\mathbf{x}\in\mathcal{X}_{\mathrm{train}} \cup \mathcal{C}$
			\State Normalize $\tilde{\mathbf{x}}\gets (\mathbf{x}-\mu_{\mathcal{C}})\oslash\sigma_{\mathcal{C}}$

			\vspace{2pt}
			\State \textbf{Train encoder $f_\theta$:}
			\State Initialize encoder parameters $\theta$ \Comment{e.g., $f_\theta(\tilde{\mathbf{x}})=W\tilde{\mathbf{x}}+b$}
			\For{each epoch}
			\For{each mini-batch $\mathcal{B}\subset \mathcal{X}_{\mathrm{train}}$}
			\State Compute latent batch $Z_{\mathcal{B}} \gets \{\mathbf{z}_i=f_\theta(\tilde{\mathbf{x}}_i)\mid \mathbf{x}_i\in \mathcal{B}\}$
			\State Compute context latents $Z_{\mathcal{C}} \gets \{\mathbf{z}=f_\theta(\tilde{\mathbf{x}})\mid \mathbf{x}\in\mathcal{C}\}$
			\State $L_{\mathrm{dep}}\gets 0$
			\For{$j=1$ to $p$}
			\State $\hat{\mu}_{ij} \gets \mathrm{TabPFNMean}(j;\, \mathbf{z}_{i,-j}, Z_{\mathcal{C}})$ for each $\mathbf{z}_i\in Z_{\mathcal{B}}$
			\State $L_{\mathrm{dep}} \gets L_{\mathrm{dep}} + \frac{1}{|\mathcal{B}|}\sum_{\mathbf{z}_i\in Z_{\mathcal{B}}}(z_{ij}-\hat{\mu}_{ij})^2$
			\EndFor
			\State $L_{\mathrm{dep}}\gets \frac{1}{p}L_{\mathrm{dep}}$
			\State $L_{\mathrm{rec}} \gets \frac{1}{|\mathcal{B}|}\sum_{\mathbf{x}_i\in\mathcal{B}} \|\tilde{\mathbf{x}}_i - W^\top \mathbf{z}_i\|_2^2$ \Comment{optional}
			\State $L_{\mathrm{enc}} \gets \lambda_{\mathrm{dep}}L_{\mathrm{dep}}+\lambda_{\mathrm{rec}}L_{\mathrm{rec}}$
			\State Update $\theta$ using $L_{\mathrm{enc}}$ (TabPFN frozen)
			\EndFor
			\EndFor
			
			\vspace{2pt}
			\State \textbf{Train variance networks $\{v_{\phi_j}\}$ on $\mathcal{X}_{\mathrm{train}}$:}
			\For{$j=1$ to $p$}
			\State Initialize $\phi_j$
			\For{each epoch}
			\For{each mini-batch $\mathcal{B}\subset \mathcal{X}_{\mathrm{train}}$}
			\State Compute $Z_{\mathcal{B}}\gets\{\mathbf{z}_i=f_\theta(\tilde{\mathbf{x}}_i)\mid \mathbf{x}_i\in\mathcal{B}\}$ and $Z_{\mathcal{C}}\gets\{\mathbf{z}=f_\theta(\tilde{\mathbf{x}})\mid \mathbf{x}\in\mathcal{C}\}$
			
			\State $\hat{\mu}_{ij} \gets \mathrm{TabPFNMean}(j;\, \mathbf{z}_{i,-j}, Z_{\mathcal{C}})$ for $\mathbf{x}_i\in\mathcal{B}$
			\State $\log \hat{\sigma}^2_{ij} \gets v_{\phi_j}(\mathbf{z}_{i,-j})$ for $\mathbf{x}_i\in\mathcal{B}$
			\State $L_{\mathrm{var}} \gets \frac{1}{|\mathcal{B}|}\sum_{\mathbf{x}_i\in\mathcal{B}}
			\left[\frac{(z_{ij}-\hat{\mu}_{ij})^2}{\hat{\sigma}^2_{ij}} + \log \hat{\sigma}^2_{ij}\right]$
			\State Update $\phi_j$ using $L_{\mathrm{var}}$ (encoder and TabPFN frozen)
			\EndFor
			\EndFor
			\EndFor
			
			\State \Return $f_\theta$, $\{v_{\phi_j}\}_{j=1}^p$, $\mathcal{C}$, $(\mu_{\mathcal{C}},\sigma_{\mathcal{C}})$
		\end{algorithmic}
	\end{algorithm}
	\noindent\textbf{Remark.} $\hat{\mu}_{ij}$ denotes the conditional mean predicted for sample $i$ and latent dimension $j$, i.e., $\hat{\mu}_{ij}\approx\mathbb{E}[Z_j\mid Z_{-j}=\mathbf{z}_{i,-j},\mathcal{C}]$. $\oslash$ denotes element-wise division.
	
	\begin{algorithm}[t]
		\caption{Inference with uLEAD-TabPFN}
		\label{alg:infer_ulead}
		\begin{algorithmic}[1]
			\Require Test sample $\mathbf{x}$, context set $\mathcal{C}$, normalization stats $(\mu_{\mathcal{C}},\sigma_{\mathcal{C}})$
			\Require Trained encoder $f_\theta$, trained variance nets $\{v_{\phi_j}\}_{j=1}^p$, frozen TabPFN
			\Ensure Anomaly score $s(\mathbf{x})$
			
			\State Normalize $\tilde{\mathbf{x}}\gets (\mathbf{x}-\mu_{\mathcal{C}})\oslash\sigma_{\mathcal{C}}$
			\State Compute $\mathbf{z} \gets f_\theta(\tilde{\mathbf{x}})$			
			\State Compute $Z_{\mathcal{C}} \gets \{f_\theta(\tilde{\mathbf{x}})\mid \tilde{\mathbf{x}}\in\tilde{\mathcal{C}}\}$
			
			\State $s(\mathbf{x})\gets 0$
			\For{$j=1$ to $p$}
			\State $\hat{\mu}_{j} \gets \mathrm{TabPFNMean}(j;\, \mathbf{z}_{-j}, Z_{\mathcal{C}})$
			\State $\log \hat{\sigma}^2_{j} \gets v_{\phi_j}(\mathbf{z}_{-j})$
			\State $s(\mathbf{x}) \gets s(\mathbf{x}) + \frac{(z_{j}-\hat{\mu}_{j})^2}{\hat{\sigma}^2_{j}}$
			\EndFor
			\State $s(\mathbf{x})\gets \frac{1}{p}s(\mathbf{x})$
			\State \Return $s(\mathbf{x})$
		\end{algorithmic}
	\end{algorithm}	
	
	\section{Appendix: Additional Results}
	\label{app:results}
	
	Tables~\ref{tab:rocaus_top10} and \ref{tab:prauc_top10} report detailed ROC-AUC and PR-AUC results for the proposed method and the top-10 baselines. Each row corresponds to one dataset and each column corresponds to one method. We report the mean ROC-AUC or PR-AUC and standard deviation over five runs (five different seeds). The value in parentheses denotes the rank of the method on that dataset (lower is better). 
	
	For the proposed method, uLEAD-TabPFN, we additionally report two relative changes: (i) the percentage change compared to the best-performing method among the full set of 25 baseline methods (denoted as \%B), and (ii) the percentage change compared to the average ROC-AUC of the 25 baseline methods (denoted as \%A). A positive value indicates an improvement and a negative value indicates a degradation. For readability, we display only the top-10 methods per dataset (based on ROC-AUC or PR-AUC) from the 25 baseline methods; however, the percentage changes are computed with respect to the full set of 25 baselines.
	
	\begin{table*}[htbp]
		\centering
		\tiny
		\setlength{\tabcolsep}{1.5pt}
		\caption{ROC-AUC comparison on ADBench datasets. We report mean$\pm$std over five runs (five different seeds). Numbers in parentheses denote the rank per dataset. uLEAD-TabPFN additionally reports relative change compared to the best baseline among the 25 baselines (\%B) and the average of the 25 baselines (\%A). For readability, only the top-10 methods (by ROC-AUC) among the 25 baseline methods are shown as columns.}
		\label{tab:rocaus_top10}
		\renewcommand{\arraystretch}{0.85}
		\resizebox{!}{0.37\textwidth}{%
			\begin{tabular}{lcccccccccc}
				\toprule
				Dataset & uLEAD-TabPFN & FoMo-0D & DTE-NP & kNN & ICL & LOF & CBLOF & FeatureBagging & SLAD & DDPM \\
				\midrule
				skin & 95.63$\pm$2.31(3)\,[{-}3.88\%B, {+}28.26\%A] & 70.60$\pm$0.88(17) & 98.86$\pm$0.46(2) & 99.49$\pm$0.08(1) & 6.58$\pm$0.63(26) & 86.34$\pm$1.77(13) & 91.82$\pm$0.21(4) & 78.39$\pm$0.91(14) & 91.05$\pm$2.61(6) & 88.74$\pm$4.40(10) \\
				smtp & 91.33$\pm$1.65(8)\,[{-}4.29\%B, {+}8.38\%A] & 83.04$\pm$4.69(18) & 92.98$\pm$2.91(5) & 92.43$\pm$2.70(6) & 74.36$\pm$7.07(23) & 93.42$\pm$2.48(4) & 87.28$\pm$5.65(12) & 84.81$\pm$3.66(15) & 92.15$\pm$1.87(7) & 95.43$\pm$1.26(1) \\
				http & 98.25$\pm$1.40(18)\,[{-}1.75\%B, {+}7.53\%A] & 99.92$\pm$0.05(10) & 99.98$\pm$0.03(4.5) & 100.00$\pm$0.00(2) & 98.24$\pm$3.45(19) & 99.98$\pm$0.03(4.5) & 99.93$\pm$0.01(9) & 92.10$\pm$0.57(21) & 99.91$\pm$0.08(11) & 100.00$\pm$0.00(2) \\
				wilt & 97.47$\pm$0.94(1)\,[{+}0.35\%B, {+}81.16\%A] & 97.13$\pm$0.00(2) & 62.91$\pm$5.59(10) & 63.66$\pm$0.00(9) & 76.42$\pm$3.46(4) & 68.81$\pm$0.00(8) & 42.90$\pm$1.14(17) & 73.35$\pm$10.50(6) & 61.76$\pm$0.11(11) & 71.66$\pm$0.81(7) \\
				vertebral & 52.78$\pm$6.69(10)\,[{-}39.47\%B, {+}2.41\%A] & 87.20$\pm$2.58(1) & 54.30$\pm$15.47(9) & 57.67$\pm$3.58(6) & 79.19$\pm$5.06(2) & 64.30$\pm$1.31(4) & 54.43$\pm$1.97(8) & 64.13$\pm$2.15(5) & 44.96$\pm$0.25(18) & 70.67$\pm$6.28(3) \\
				annthyroid & 81.60$\pm$5.96(17)\,[{-}12.58\%B, {-}2.22\%A] & 85.17$\pm$0.00(16) & 92.90$\pm$0.25(3) & 92.81$\pm$0.00(4) & 81.11$\pm$1.07(18) & 88.63$\pm$0.00(11) & 90.14$\pm$1.08(7) & 88.95$\pm$1.61(8) & 93.34$\pm$0.36(1) & 88.82$\pm$1.34(10) \\
				mammography & 71.73$\pm$2.77(23)\,[{-}20.89\%B, {-}12.56\%A] & 69.11$\pm$1.21(26) & 87.62$\pm$0.09(8) & 87.58$\pm$0.00(9) & 71.87$\pm$9.11(22) & 85.52$\pm$0.00(12) & 84.74$\pm$0.02(14) & 86.31$\pm$0.35(10) & 74.51$\pm$0.67(19) & 81.01$\pm$2.04(16) \\
				thyroid & 88.18$\pm$3.89(26)\,[{-}10.89\%B, {-}8.36\%A] & 96.84$\pm$0.00(13) & 98.63$\pm$0.04(4) & 98.68$\pm$0.00(2) & 95.40$\pm$0.98(16) & 92.72$\pm$0.00(23) & 98.54$\pm$0.06(8) & 93.15$\pm$1.52(22) & 95.27$\pm$0.24(17) & 97.95$\pm$0.22(11) \\
				glass & 87.59$\pm$2.42(9)\,[{-}11.91\%B, {+}9.12\%A] & 98.30$\pm$0.53(2) & 89.64$\pm$3.54(5) & 92.04$\pm$1.12(4) & 99.44$\pm$0.58(1) & 88.82$\pm$1.98(7) & 89.35$\pm$1.48(6) & 88.49$\pm$1.71(8) & 86.04$\pm$5.69(10) & 66.67$\pm$13.63(23) \\
				yeast & 57.41$\pm$1.07(1)\,[{+}9.28\%B, {+}25.06\%A] & 49.91$\pm$0.00(5) & 44.58$\pm$0.33(19) & 44.74$\pm$0.00(17) & 48.98$\pm$2.39(7) & 45.79$\pm$0.00(14) & 50.40$\pm$0.08(4) & 46.40$\pm$1.33(13) & 48.69$\pm$0.11(8) & 49.13$\pm$2.85(6) \\
				pima & 64.75$\pm$3.46(17)\,[{-}20.55\%B, {-}5.03\%A] & 80.46$\pm$1.48(2) & 81.50$\pm$2.57(1) & 76.94$\pm$1.87(4) & 79.68$\pm$3.06(3) & 70.53$\pm$2.19(14) & 72.94$\pm$1.07(9) & 71.93$\pm$2.22(12) & 60.62$\pm$4.21(20) & 70.27$\pm$2.28(15) \\
				stamps & 84.79$\pm$3.37(20)\,[{-}13.37\%B, {-}3.08\%A] & 97.52$\pm$0.39(2) & 97.87$\pm$0.37(1) & 95.89$\pm$1.44(4) & 96.68$\pm$1.09(3) & 93.74$\pm$2.36(6) & 93.41$\pm$1.66(9) & 94.21$\pm$2.26(5) & 81.97$\pm$8.29(21) & 91.84$\pm$4.15(14) \\
				wbc & 79.98$\pm$5.19(24)\,[{-}19.84\%B, {-}14.44\%A] & 99.61$\pm$0.31(4) & 99.54$\pm$0.28(5) & 99.12$\pm$0.19(13) & 99.66$\pm$0.36(2) & 80.52$\pm$4.47(23) & 98.31$\pm$1.31(16) & 58.05$\pm$18.08(25) & 99.78$\pm$0.18(1) & 99.20$\pm$0.39(11) \\
				breastw & 87.14$\pm$1.20(23)\,[{-}12.45\%B, {-}5.89\%A] & 99.15$\pm$0.23(9) & 99.28$\pm$0.12(5) & 99.05$\pm$0.26(12) & 98.28$\pm$0.45(16) & 88.91$\pm$6.80(22) & 99.11$\pm$0.23(11) & 59.07$\pm$15.44(24) & 99.53$\pm$0.13(1) & 98.70$\pm$0.43(14) \\
				shuttle & 99.23$\pm$0.12(17)\,[{-}0.75\%B, {+}5.99\%A] & 99.82$\pm$0.09(7) & 99.93$\pm$0.02(2) & 99.91$\pm$0.00(4.5) & 99.92$\pm$0.04(3) & 99.98$\pm$0.00(1) & 99.72$\pm$0.02(10) & 86.89$\pm$8.22(21) & 99.90$\pm$0.01(6) & 99.91$\pm$0.01(4.5) \\
				magic.gamma & 83.46$\pm$1.09(5)\,[{-}2.91\%B, {+}13.46\%A] & 84.77$\pm$0.14(2) & 83.57$\pm$0.76(4) & 83.27$\pm$0.00(7) & 75.56$\pm$0.42(11) & 83.40$\pm$0.00(6) & 75.81$\pm$0.00(10) & 84.19$\pm$0.72(3) & 72.00$\pm$0.01(16) & 85.97$\pm$1.08(1) \\
				pageblocks & 93.45$\pm$1.51(1)\,[{+}1.22\%B, {+}9.30\%A] & 86.58$\pm$0.00(16) & 89.32$\pm$0.26(7) & 89.65$\pm$0.00(6) & 88.39$\pm$0.77(9) & 91.30$\pm$0.00(3) & 91.23$\pm$0.11(4) & 91.11$\pm$0.30(5) & 87.86$\pm$0.01(12) & 86.93$\pm$0.42(15) \\
				donors & 99.99$\pm$0.01(1)\,[{+}0.08\%B, {+}18.46\%A] & 99.91$\pm$0.02(2) & 99.26$\pm$0.29(5) & 99.49$\pm$0.06(4) & 99.90$\pm$0.05(3) & 96.97$\pm$0.24(6) & 93.47$\pm$0.23(8) & 95.21$\pm$1.72(7) & 88.51$\pm$5.51(15) & 82.50$\pm$1.83(17) \\
				cover & 82.04$\pm$5.04(17)\,[{-}17.28\%B, {-}0.05\%A] & 99.12$\pm$0.29(3) & 97.73$\pm$0.55(5) & 97.54$\pm$0.15(6) & 89.34$\pm$4.02(14) & 99.18$\pm$0.10(1) & 94.04$\pm$0.28(12) & 99.16$\pm$0.63(2) & 73.97$\pm$13.80(20) & 98.35$\pm$0.66(4) \\
				vowels & 97.97$\pm$1.01(1)\,[{+}10.72\%B, {+}46.45\%A] & 81.29$\pm$0.00(10) & 81.42$\pm$1.44(9) & 82.21$\pm$0.00(8) & 85.10$\pm$2.13(6) & 86.30$\pm$0.00(4) & 78.72$\pm$4.01(11) & 85.32$\pm$1.16(5) & 85.02$\pm$0.02(7) & 86.38$\pm$1.94(3) \\
				wine & 95.73$\pm$5.14(12)\,[{-}4.27\%B, {+}5.29\%A] & 99.94$\pm$0.12(2) & 99.44$\pm$0.97(5) & 99.19$\pm$0.22(6) & 99.87$\pm$0.29(3) & 98.36$\pm$0.38(7) & 97.75$\pm$0.59(10) & 97.93$\pm$0.87(8) & 100.00$\pm$0.00(1) & 99.61$\pm$0.09(4) \\
				pendigits & 97.91$\pm$1.20(7)\,[{-}1.97\%B, {+}9.88\%A] & 98.11$\pm$0.00(5.5) & 99.61$\pm$0.17(2) & 99.87$\pm$0.00(1) & 96.71$\pm$0.83(9) & 99.05$\pm$0.00(4) & 96.67$\pm$0.07(10) & 99.50$\pm$0.11(3) & 94.61$\pm$0.41(12) & 98.11$\pm$0.23(5.5) \\
				lymphography & 88.08$\pm$9.11(24)\,[{-}11.92\%B, {-}7.63\%A] & 99.93$\pm$0.11(6) & 99.93$\pm$0.09(6) & 99.93$\pm$0.08(6) & 100.00$\pm$0.00(2) & 98.21$\pm$0.75(21) & 99.83$\pm$0.02(12) & 96.61$\pm$2.84(22) & 100.00$\pm$0.01(2) & 99.94$\pm$0.09(4) \\
				hepatitis & 75.73$\pm$3.07(20)\,[{-}24.22\%B, {-}10.91\%A] & 99.88$\pm$0.24(3) & 93.22$\pm$3.90(9) & 96.46$\pm$1.46(6) & 99.94$\pm$0.13(1) & 66.92$\pm$7.01(25) & 86.26$\pm$2.35(12) & 67.76$\pm$6.50(24) & 99.93$\pm$0.15(2) & 97.74$\pm$1.18(5) \\
				cardiotocography & 71.07$\pm$3.69(9)\,[{-}10.38\%B, {+}10.68\%A] & 72.15$\pm$0.00(8) & 63.76$\pm$1.88(15) & 62.11$\pm$0.00(18) & 54.20$\pm$1.80(22) & 64.49$\pm$0.00(14) & 67.61$\pm$2.21(11) & 63.64$\pm$2.40(16) & 47.32$\pm$0.31(24) & 54.54$\pm$2.96(21) \\
				waveform & 62.00$\pm$5.28(19)\,[{-}19.43\%B, {-}6.34\%A] & 71.02$\pm$0.00(9) & 74.48$\pm$0.00(5) & 75.21$\pm$0.00(4) & 68.68$\pm$3.66(12) & 76.00$\pm$0.00(2) & 72.93$\pm$0.90(6) & 76.95$\pm$1.07(1) & 48.92$\pm$0.08(26) & 62.17$\pm$2.64(18) \\
				cardio & 84.84$\pm$1.39(18)\,[{-}12.13\%B, {-}2.51\%A] & 94.27$\pm$0.00(6) & 91.80$\pm$0.59(13) & 92.00$\pm$0.00(12) & 80.01$\pm$2.12(22) & 92.21$\pm$0.00(10) & 93.49$\pm$1.47(7) & 92.12$\pm$0.56(11) & 83.05$\pm$1.10(19) & 86.94$\pm$1.96(16) \\
				fault & 75.13$\pm$0.76(1)\,[{+}17.51\%B, {+}33.41\%A] & 61.82$\pm$0.00(4) & 58.64$\pm$0.65(12) & 58.73$\pm$0.00(11) & 60.63$\pm$0.50(6) & 47.42$\pm$0.00(26) & 59.00$\pm$1.29(9) & 48.32$\pm$0.95(25) & 63.93$\pm$0.19(2) & 61.09$\pm$1.16(5) \\
				aloi & 58.47$\pm$0.56(1)\,[{+}7.70\%B, {+}14.77\%A] & 54.05$\pm$0.30(3) & 51.19$\pm$0.46(10) & 51.04$\pm$0.00(12) & 47.50$\pm$0.98(26) & 48.76$\pm$0.00(22) & 53.69$\pm$0.15(7) & 49.07$\pm$0.53(21) & 50.76$\pm$0.29(15) & 49.91$\pm$0.35(18) \\
				fraud & 95.41$\pm$0.61(5)\,[{-}0.24\%B, {+}6.00\%A] & 93.91$\pm$2.81(15) & 95.64$\pm$1.05(1) & 95.43$\pm$1.04(4) & 92.78$\pm$1.46(18) & 94.35$\pm$1.41(13) & 94.91$\pm$1.08(8) & 94.83$\pm$1.56(10) & 94.38$\pm$1.01(12) & 93.65$\pm$0.92(16) \\
				wdbc & 97.77$\pm$1.17(19)\,[{-}2.02\%B, {+}3.79\%A] & 99.77$\pm$0.25(2) & 99.52$\pm$0.38(5) & 99.05$\pm$0.27(13) & 99.78$\pm$0.27(1) & 99.62$\pm$0.21(4) & 98.73$\pm$0.43(16.5) & 99.63$\pm$0.18(3) & 99.49$\pm$0.25(6) & 99.30$\pm$0.17(9) \\
				letter & 93.24$\pm$1.60(1)\,[{+}25.88\%B, {+}141.83\%A] & 40.90$\pm$0.00(8) & 34.38$\pm$0.98(17) & 35.43$\pm$0.00(16) & 42.68$\pm$1.17(7) & 44.83$\pm$0.00(6) & 33.24$\pm$0.66(19) & 44.84$\pm$1.00(5) & 36.80$\pm$0.41(12) & 38.05$\pm$1.12(11) \\
				ionosphere & 95.62$\pm$1.16(10)\,[{-}3.40\%B, {+}6.46\%A] & 96.59$\pm$1.25(8) & 97.77$\pm$1.39(3) & 97.44$\pm$0.98(4) & 98.98$\pm$0.32(1) & 94.29$\pm$2.20(14) & 96.78$\pm$1.50(7) & 94.47$\pm$2.10(13) & 98.21$\pm$0.60(2) & 94.60$\pm$0.85(12) \\
				wpbc & 55.76$\pm$1.97(17)\,[{-}42.29\%B, {-}11.46\%A] & 76.10$\pm$2.11(6) & 83.16$\pm$13.46(3) & 63.67$\pm$2.34(8) & 96.61$\pm$1.17(1) & 57.36$\pm$2.01(14) & 59.57$\pm$1.98(12) & 56.79$\pm$1.86(15) & 95.53$\pm$2.21(2) & 66.49$\pm$3.17(7) \\
				satellite & 82.72$\pm$0.57(5)\,[{-}5.45\%B, {+}9.17\%A] & 84.98$\pm$0.00(4) & 82.11$\pm$0.67(7) & 82.24$\pm$0.00(6) & 85.15$\pm$0.48(3) & 80.30$\pm$0.00(8) & 73.20$\pm$0.92(17) & 80.11$\pm$0.11(9) & 87.49$\pm$0.10(1) & 77.70$\pm$0.53(11) \\
				satimage-2 & 99.66$\pm$0.30(5)\,[{-}0.26\%B, {+}1.84\%A] & 98.40$\pm$0.00(17) & 99.67$\pm$0.00(4) & 99.71$\pm$0.00(3) & 99.48$\pm$0.22(8) & 99.38$\pm$0.00(11) & 99.42$\pm$0.01(10) & 99.47$\pm$0.03(9) & 99.77$\pm$0.00(2) & 99.62$\pm$0.16(6) \\
				landsat & 67.32$\pm$0.52(5)\,[{-}8.05\%B, {+}18.41\%A] & 69.53$\pm$0.00(2) & 68.20$\pm$1.75(4) & 68.25$\pm$0.00(3) & 65.13$\pm$0.44(8) & 66.58$\pm$0.00(6) & 57.21$\pm$0.26(12) & 66.38$\pm$0.17(7) & 65.03$\pm$0.18(9) & 51.37$\pm$1.00(19) \\
				celeba & 74.43$\pm$0.94(11)\,[{-}11.78\%B, {+}8.36\%A] & 74.81$\pm$0.61(10) & 70.40$\pm$0.37(16) & 73.14$\pm$0.72(12) & 72.21$\pm$0.82(13) & 43.73$\pm$0.75(26) & 79.28$\pm$1.30(5) & 46.89$\pm$1.80(24) & 67.43$\pm$1.64(18) & 78.56$\pm$1.99(6) \\
				spambase & 80.48$\pm$1.07(14)\,[{-}5.52\%B, {+}5.50\%A] & 78.21$\pm$0.00(15) & 83.74$\pm$0.69(3) & 83.36$\pm$0.00(5) & 83.53$\pm$0.45(4) & 73.23$\pm$0.00(18) & 81.52$\pm$0.55(10) & 69.64$\pm$2.09(22) & 84.86$\pm$0.20(2) & 64.54$\pm$0.80(25) \\
				campaign & 74.71$\pm$1.11(13)\,[{-}7.68\%B, {+}5.83\%A] & 65.30$\pm$0.70(20) & 78.79$\pm$0.25(2) & 78.48$\pm$0.00(4) & 80.92$\pm$0.79(1) & 70.55$\pm$0.00(16) & 77.05$\pm$0.31(10) & 69.10$\pm$3.85(19) & 76.75$\pm$0.16(12) & 74.51$\pm$0.42(14) \\
				optdigits & 85.48$\pm$5.00(9)\,[{-}12.04\%B, {+}20.69\%A] & 85.28$\pm$0.00(10) & 94.28$\pm$1.67(5) & 93.72$\pm$0.00(6) & 97.18$\pm$0.80(1) & 96.65$\pm$0.00(2) & 83.52$\pm$1.69(12) & 96.27$\pm$0.49(3) & 95.28$\pm$0.17(4) & 90.76$\pm$2.12(7) \\
				mnist & 88.30$\pm$1.60(13.5)\,[{-}6.09\%B, {+}9.41\%A] & 91.27$\pm$0.00(5) & 94.02$\pm$0.42(1) & 93.85$\pm$0.00(2) & 90.11$\pm$1.13(10) & 92.93$\pm$0.00(3) & 91.10$\pm$0.23(6) & 92.55$\pm$0.40(4) & 89.73$\pm$0.44(12) & 87.27$\pm$3.22(15) \\
				musk & 100.00$\pm$0.00(7.5)\,[{+}0.00\%B, {+}4.82\%A] & 99.62$\pm$0.43(19) & 100.00$\pm$0.00(7.5) & 100.00$\pm$0.00(7.5) & 99.37$\pm$0.63(20) & 100.00$\pm$0.00(7.5) & 100.00$\pm$0.00(7.5) & 100.00$\pm$0.00(7.5) & 100.00$\pm$0.00(7.5) & 100.00$\pm$0.00(7.5) \\
				backdoor & 96.07$\pm$0.46(1)\,[{+}0.77\%B, {+}29.13\%A] & 78.59$\pm$17.11(13) & 93.31$\pm$1.70(7) & 93.75$\pm$0.53(5) & 93.62$\pm$0.69(6) & 95.33$\pm$0.25(2) & 69.65$\pm$5.23(17) & 94.83$\pm$0.62(3) & 50.00$\pm$0.00(24) & 80.93$\pm$0.56(12) \\
				speech & 60.53$\pm$2.26(1)\,[{+}15.76\%B, {+}49.22\%A] & 40.10$\pm$1.75(10) & 41.37$\pm$0.00(8.5) & 36.36$\pm$0.00(24) & 48.86$\pm$2.72(6) & 37.53$\pm$0.00(15) & 35.88$\pm$0.15(26) & 37.48$\pm$0.37(16) & 41.37$\pm$0.94(8.5) & 36.96$\pm$0.86(18) \\
				census & 70.49$\pm$1.23(8)\,[{-}4.93\%B, {+}14.73\%A] & 60.54$\pm$4.96(15) & 72.10$\pm$0.40(3) & 72.26$\pm$0.29(2) & 70.56$\pm$0.35(5) & 58.46$\pm$1.06(17) & 70.84$\pm$0.28(4) & 55.92$\pm$1.00(19) & 57.91$\pm$10.84(18) & 70.15$\pm$0.23(9) \\
				CIFAR10 & 62.16$\pm$7.07(18)\,[{-}11.59\%B, {-}1.58\%A] & 64.53$\pm$0.62(14) & 67.82$\pm$0.00(5) & 67.53$\pm$0.00(7.5) & 63.60$\pm$0.83(16) & 70.30$\pm$0.00(2) & 67.87$\pm$0.21(4) & 70.31$\pm$0.21(1) & 66.60$\pm$0.10(12) & 67.91$\pm$0.13(3) \\
				SVHN & 59.79$\pm$2.64(15)\,[{-}6.48\%B, {+}2.56\%A] & 59.93$\pm$0.64(14) & 62.13$\pm$0.00(3) & 61.69$\pm$0.00(5) & 61.70$\pm$0.50(4) & 63.82$\pm$0.00(2) & 60.97$\pm$0.17(9) & 63.93$\pm$0.14(1) & 60.87$\pm$0.07(10) & 61.37$\pm$0.08(6) \\
				MVTec-AD & 78.31$\pm$10.98(12)\,[{-}17.36\%B, {+}1.78\%A] & 75.31$\pm$1.26(20) & 89.66$\pm$1.51(3) & 81.51$\pm$1.82(7) & 94.75$\pm$0.84(1) & 80.36$\pm$2.14(10) & 79.96$\pm$2.02(11) & 80.45$\pm$2.20(9) & 93.24$\pm$1.09(2) & 78.00$\pm$1.99(13) \\
				FashionMNIST & 88.62$\pm$6.53(9)\,[{-}3.33\%B, {+}9.80\%A] & 86.66$\pm$0.66(15) & 90.14$\pm$0.00(4) & 89.87$\pm$0.00(6) & 90.56$\pm$0.27(3) & 91.60$\pm$0.00(2) & 89.10$\pm$0.15(8) & 91.67$\pm$0.11(1) & 89.94$\pm$0.04(5) & 88.52$\pm$0.07(10) \\
				MNIST-C & 86.71$\pm$11.46(3)\,[{-}0.70\%B, {+}17.22\%A] & 77.28$\pm$1.69(15) & 84.74$\pm$0.00(5) & 84.11$\pm$0.00(6) & 85.62$\pm$0.52(4) & 87.21$\pm$0.00(2) & 81.11$\pm$0.09(8) & 87.33$\pm$0.15(1) & 83.51$\pm$0.11(7) & 80.12$\pm$0.14(9) \\
				yelp & 68.18$\pm$0.43(2)\,[{-}0.70\%B, {+}15.47\%A] & 60.29$\pm$5.14(11) & 68.66$\pm$0.00(1) & 68.07$\pm$0.00(3) & 55.79$\pm$0.50(20) & 67.20$\pm$0.00(4) & 63.80$\pm$0.07(7) & 67.06$\pm$0.12(5) & 54.88$\pm$0.19(21) & 59.30$\pm$0.09(14) \\
				imdb & 52.25$\pm$0.73(2)\,[{-}0.17\%B, {+}5.09\%A] & 50.84$\pm$6.24(9) & 50.43$\pm$0.00(10) & 50.08$\pm$0.00(11) & 52.34$\pm$0.49(1) & 49.57$\pm$0.00(15) & 49.94$\pm$0.01(13.5) & 49.53$\pm$0.10(16.5) & 51.26$\pm$0.10(6) & 47.91$\pm$0.10(24) \\
				agnews & 72.44$\pm$5.52(3)\,[{-}2.95\%B, {+}22.23\%A] & 60.96$\pm$3.69(9) & 67.98$\pm$0.00(4.5) & 67.05$\pm$0.00(6) & 62.55$\pm$0.47(8) & 74.58$\pm$0.00(2) & 62.82$\pm$0.08(7) & 74.64$\pm$0.09(1) & 58.43$\pm$0.07(13.5) & 57.82$\pm$0.12(15) \\
				20news & 61.37$\pm$9.36(2)\,[{-}2.35\%B, {+}9.30\%A] & 56.93$\pm$1.46(11) & 59.97$\pm$0.74(6) & 57.36$\pm$0.66(9) & 61.30$\pm$1.05(3) & 60.19$\pm$0.62(5) & 57.10$\pm$1.23(10) & 60.25$\pm$0.70(4) & 59.47$\pm$0.80(8) & 54.88$\pm$0.52(16) \\
				amazon & 62.01$\pm$0.41(1)\,[{+}1.95\%B, {+}16.80\%A] & 54.74$\pm$4.18(15) & 60.82$\pm$0.00(2) & 60.58$\pm$0.00(3) & 54.21$\pm$0.22(16) & 57.88$\pm$0.00(7) & 58.17$\pm$0.12(5) & 57.94$\pm$0.04(6) & 52.01$\pm$0.06(20) & 55.09$\pm$0.10(13) \\
				internetads & 90.76$\pm$0.32(1)\,[{+}19.51\%B, {+}43.72\%A] & 55.64$\pm$7.12(20) & 69.96$\pm$2.22(8) & 68.08$\pm$0.00(10) & 72.20$\pm$0.55(4) & 71.72$\pm$0.00(5) & 65.16$\pm$0.08(16) & 71.38$\pm$2.32(6) & 75.94$\pm$0.11(2) & 65.76$\pm$0.06(13) \\
				\bottomrule
			\end{tabular}%
		}
		
	\end{table*}
	
	% PR-AUC Top-10
	\begin{table*}[htbp]
		\centering
		\tiny
		\caption{PR-AUC comparison on ADBench datasets. We report mean$\pm$std over five runs (five different seeds). Numbers in parentheses denote the rank per dataset. uLEAD-TabPFN additionally reports relative change compared to the best baseline among the 25 baselines (\%B) and the average of the 25 baselines (\%A). For readability, only the top-10 methods (by PR-AUC) among the 25 baseline methods are shown as columns.}
		\label{tab:prauc_top10}
		\setlength{\tabcolsep}{1.5pt}
		\renewcommand{\arraystretch}{0.85}
		\resizebox{!}{0.37\textwidth}{%
			\begin{tabular}{lcccccccccc}
				\toprule
				Dataset & uLEAD-TabPFN & DDAE & FoMo-0D & DTE-NP & kNN & ICL & LOF & SLAD & DDPM & OCSVM \\
				\midrule
				skin & 86.77$\pm$8.03(4) [-11.67\%B, +49.27\%A] & 92.23$\pm$0.01(3) & 51.04$\pm$1.72(17) & 94.78$\pm$2.33(2) & 98.24$\pm$0.41(1) & 32.46$\pm$1.00(24) & 61.68$\pm$1.85(14) & 78.73$\pm$7.88(5) & 76.37$\pm$5.79(6) & 66.31$\pm$0.51(9) \\
				smtp & 33.48$\pm$6.18(14) [-50.78\%B, +13.69\%A] & 46.23$\pm$0.08(11) & 38.18$\pm$9.90(13) & 50.20$\pm$6.39(5) & 50.53$\pm$5.92(4) & 3.81$\pm$3.83(20) & 48.09$\pm$7.38(10) & 50.60$\pm$6.30(3) & 40.81$\pm$13.36(12) & 64.51$\pm$11.88(2) \\
				http & 48.70$\pm$1.46(19) [-51.30\%B, -23.81\%A] & 99.25$\pm$0.01(4) & 90.58$\pm$5.01(9) & 97.10$\pm$4.04(6) & 100.00$\pm$0.00(1) & 70.82$\pm$42.06(13) & 97.12$\pm$3.92(5) & 88.09$\pm$7.46(12) & 99.96$\pm$0.06(2) & 99.88$\pm$0.26(3) \\
				wilt & 84.63$\pm$5.63(1) [+9.26\%B, +490.69\%A] & 14.79$\pm$0.02(9) & 77.46$\pm$0.00(2) & 12.20$\pm$1.60(11) & 12.25$\pm$0.00(10) & 28.94$\pm$3.31(3) & 15.74$\pm$0.00(8) & 12.17$\pm$0.04(12) & 17.24$\pm$0.46(6) & 7.12$\pm$0.00(24) \\
				vertebral & 22.33$\pm$2.84(16) [-67.83\%B, -16.50\%A] & 27.75$\pm$0.04(7) & 69.39$\pm$3.97(1) & 25.21$\pm$8.90(10) & 26.11$\pm$2.49(8) & 58.75$\pm$7.30(2) & 33.87$\pm$4.30(4) & 19.87$\pm$4.37(22) & 35.84$\pm$9.27(3) & 22.23$\pm$2.01(17) \\
				annthyroid & 45.90$\pm$3.90(21) [-34.97\%B, -14.48\%A] & 56.88$\pm$0.02(13) & 48.81$\pm$0.00(18) & 68.15$\pm$0.38(2) & 68.07$\pm$0.00(3) & 45.83$\pm$2.16(22) & 53.53$\pm$0.00(16) & 70.58$\pm$0.92(1) & 62.88$\pm$2.59(7) & 60.11$\pm$0.00(9) \\
				mammography & 28.64$\pm$3.46(16) [-48.11\%B, -10.38\%A] & 35.21$\pm$0.07(13) & 36.28$\pm$1.26(12) & 42.09$\pm$0.86(4) & 41.27$\pm$0.00(7) & 17.11$\pm$3.75(25) & 34.07$\pm$0.00(14) & 18.98$\pm$1.05(23) & 19.93$\pm$3.92(22) & 40.52$\pm$0.00(9) \\
				thyroid & 56.84$\pm$9.20(23) [-30.86\%B, -17.73\%A] & 67.34$\pm$0.03(16) & 67.00$\pm$0.00(17) & 81.03$\pm$0.31(5) & 80.94$\pm$0.00(6) & 51.51$\pm$12.75(25) & 60.57$\pm$0.00(22) & 74.07$\pm$0.84(14) & 82.22$\pm$0.87(1) & 78.92$\pm$0.00(10) \\
				glass & 29.61$\pm$8.62(14) [-67.94\%B, -14.27\%A] & 31.47$\pm$0.07(11) & 83.60$\pm$4.63(2) & 37.38$\pm$15.51(8) & 42.32$\pm$8.47(5) & 92.35$\pm$8.32(1) & 38.12$\pm$9.91(7) & 41.15$\pm$3.98(6) & 31.21$\pm$11.30(12) & 26.76$\pm$7.68(16) \\
				yeast & 55.76$\pm$1.16(2) [-17.77\%B, +12.72\%A] & 67.81$\pm$0.02(1) & 51.00$\pm$0.00(5) & 48.12$\pm$0.49(19) & 48.26$\pm$0.00(18) & 49.55$\pm$1.36(12) & 48.94$\pm$0.00(17) & 50.56$\pm$0.08(8) & 51.05$\pm$1.65(4) & 47.95$\pm$0.00(20) \\
				pima & 65.24$\pm$4.65(18) [-20.91\%B, -5.24\%A] & 82.49$\pm$0.02(1) & 79.27$\pm$1.74(3) & 79.70$\pm$2.44(2) & 75.37$\pm$2.99(6) & 78.63$\pm$1.94(4) & 68.40$\pm$3.79(17) & 63.03$\pm$4.45(21) & 71.18$\pm$2.06(12.5) & 71.98$\pm$3.41(9) \\
				stamps & 51.52$\pm$9.84(18) [-42.35\%B, -11.07\%A] & 73.81$\pm$0.12(4) & 89.37$\pm$3.50(1) & 82.47$\pm$4.08(2) & 71.68$\pm$8.35(5) & 79.54$\pm$5.44(3) & 64.84$\pm$8.17(8) & 50.61$\pm$13.26(19) & 64.74$\pm$12.87(9) & 64.91$\pm$7.95(7) \\
				wbc & 53.93$\pm$5.84(24) [-45.04\%B, -31.80\%A] & 87.83$\pm$0.03(15) & 96.46$\pm$2.49(3) & 96.10$\pm$2.17(4) & 92.01$\pm$3.96(12) & 95.12$\pm$5.14(5) & 24.89$\pm$3.49(25) & 98.14$\pm$1.35(1) & 93.84$\pm$2.45(8) & 97.15$\pm$0.77(2) \\
				breastw & 79.90$\pm$3.24(24) [-19.70\%B, -13.62\%A] & 99.51$\pm$0.00(1) & 99.03$\pm$0.34(12) & 99.19$\pm$0.14(6.5) & 98.92$\pm$0.32(13) & 96.79$\pm$1.61(18) & 80.01$\pm$10.09(23) & 99.47$\pm$0.21(3) & 98.60$\pm$0.51(15) & 99.35$\pm$0.21(5) \\
				shuttle & 98.53$\pm$0.31(6) [-1.23\%B, +14.50\%A] & 99.11$\pm$0.00(4) & 99.36$\pm$0.17(3) & 98.14$\pm$0.48(7) & 97.86$\pm$0.00(12) & 99.72$\pm$0.14(2) & 99.75$\pm$0.00(1) & 98.04$\pm$0.01(9) & 97.91$\pm$0.26(11) & 97.67$\pm$0.00(13) \\
				magic.gamma & 85.62$\pm$1.41(8) [-8.50\%B, +9.28\%A] & 93.57$\pm$0.00(1) & 87.54$\pm$0.18(3) & 86.15$\pm$0.74(6) & 85.86$\pm$0.00(7) & 81.33$\pm$0.57(10) & 86.36$\pm$0.00(5) & 77.34$\pm$0.01(15) & 87.97$\pm$0.84(2) & 79.16$\pm$0.00(13) \\
				pageblocks & 81.89$\pm$3.04(2) [-3.70\%B, +36.23\%A] & 85.04$\pm$0.01(1) & 62.67$\pm$0.00(14) & 67.45$\pm$0.10(9) & 67.60$\pm$0.00(8) & 68.11$\pm$2.30(7) & 71.07$\pm$0.00(4) & 64.70$\pm$0.79(10) & 62.10$\pm$0.95(15) & 64.25$\pm$0.00(11) \\
				donors & 99.88$\pm$0.11(1) [+1.17\%B, +107.39\%A] & 98.73$\pm$0.01(2) & 98.43$\pm$0.48(3) & 85.55$\pm$4.56(6) & 89.09$\pm$0.94(5) & 98.35$\pm$0.87(4) & 63.39$\pm$1.89(8) & 46.18$\pm$9.80(11) & 26.66$\pm$2.91(23) & 42.71$\pm$0.86(13) \\
				cover & 36.55$\pm$4.25(8) [-55.93\%B, +25.11\%A] & 79.50$\pm$0.02(2) & 72.57$\pm$8.38(5) & 59.97$\pm$10.57(6) & 55.79$\pm$3.74(7) & 34.48$\pm$16.39(9) & 82.92$\pm$2.19(1) & 6.96$\pm$5.18(21) & 73.27$\pm$3.36(4) & 22.28$\pm$1.01(13) \\
				vowels & 79.51$\pm$8.49(2) [-15.99\%B, +233.47\%A] & 94.64$\pm$0.02(1) & 24.10$\pm$0.00(12) & 31.59$\pm$1.59(8) & 30.21$\pm$0.00(9) & 27.39$\pm$5.75(11) & 33.09$\pm$0.00(6) & 39.23$\pm$1.65(5) & 42.72$\pm$4.80(4) & 27.43$\pm$0.00(10) \\
				wine & 82.77$\pm$18.44(12) [-17.23\%B, +11.62\%A] & 79.15$\pm$0.06(13) & 99.62$\pm$0.77(2) & 96.80$\pm$5.58(5) & 95.11$\pm$1.81(6) & 98.26$\pm$3.89(3) & 89.95$\pm$2.64(7) & 100.00$\pm$0.00(1) & 97.65$\pm$0.72(4) & 88.68$\pm$2.29(8.5) \\
				pendigits & 71.43$\pm$11.94(6) [-26.35\%B, +54.77\%A] & 92.65$\pm$0.01(2) & 66.33$\pm$0.00(8) & 91.89$\pm$3.30(3) & 96.99$\pm$0.00(1) & 66.41$\pm$7.58(7) & 78.55$\pm$0.00(5) & 35.35$\pm$0.15(18) & 61.14$\pm$4.73(9) & 51.78$\pm$0.00(11) \\
				lymphography & 66.31$\pm$19.50(25) [-33.69\%B, -25.56\%A] & 92.52$\pm$0.08(19) & 99.15$\pm$1.26(7) & 99.34$\pm$0.93(3.5) & 99.17$\pm$0.94(6) & 100.00$\pm$0.00(1.5) & 84.16$\pm$4.27(22) & 99.34$\pm$0.12(3.5) & 99.30$\pm$1.02(5) & 100.00$\pm$0.00(1.5) \\
				hepatitis & 58.35$\pm$5.37(17) [-41.55\%B, -16.36\%A] & 54.92$\pm$0.07(21) & 99.45$\pm$1.10(3) & 82.32$\pm$10.26(9) & 90.31$\pm$4.36(6) & 99.83$\pm$0.38(1) & 43.67$\pm$10.79(26) & 99.79$\pm$0.47(2) & 95.14$\pm$1.34(5) & 77.63$\pm$3.49(10) \\
				cardiotocography & 63.69$\pm$2.92(7) [-8.61\%B, +10.74\%A] & 68.19$\pm$0.02(4) & 63.55$\pm$0.00(8) & 58.68$\pm$1.14(14) & 57.43$\pm$0.00(15) & 48.66$\pm$3.84(24) & 57.32$\pm$0.00(16) & 49.37$\pm$0.19(23) & 51.31$\pm$1.98(21) & 66.19$\pm$0.00(6) \\
				waveform & 20.36$\pm$4.07(8) [-33.59\%B, +42.43\%A] & 21.00$\pm$0.04(7) & 9.95$\pm$0.00(15) & 27.87$\pm$0.00(3) & 27.00$\pm$0.00(4) & 18.63$\pm$4.08(10) & 30.66$\pm$0.00(1) & 5.31$\pm$0.01(27) & 9.31$\pm$1.05(17) & 10.91$\pm$0.00(13) \\
				cardio & 64.76$\pm$5.37(21) [-24.92\%B, -6.77\%A] & 79.60$\pm$0.02(6) & 78.92$\pm$0.00(7) & 77.41$\pm$0.85(10) & 77.22$\pm$0.00(11) & 47.91$\pm$11.47(25) & 70.15$\pm$0.00(15) & 69.94$\pm$0.98(16) & 69.28$\pm$0.91(17) & 83.59$\pm$0.00(4) \\
				fault & 73.05$\pm$0.79(2) [-17.49\%B, +21.25\%A] & 88.53$\pm$0.01(1) & 63.08$\pm$0.00(7) & 62.17$\pm$0.14(9) & 61.98$\pm$0.00(11) & 63.18$\pm$0.70(6) & 50.44$\pm$0.00(27) & 66.69$\pm$0.19(3) & 64.75$\pm$0.69(4) & 61.12$\pm$0.00(13) \\
				aloi & 10.39$\pm$0.41(1) [+19.31\%B, +65.46\%A] & 8.71$\pm$0.00(2) & 6.93$\pm$0.05(4) & 6.05$\pm$0.06(15) & 6.02$\pm$0.00(16) & 5.50$\pm$0.13(26) & 6.54$\pm$0.00(7) & 5.99$\pm$0.04(17) & 5.97$\pm$0.06(18) & 6.52$\pm$0.00(9) \\
				fraud & 71.63$\pm$5.21(1) [+3.49\%B, +82.40\%A] & 48.24$\pm$0.04(11) & 55.13$\pm$13.77(7) & 42.10$\pm$7.63(13) & 38.68$\pm$7.19(14) & 53.88$\pm$8.78(9) & 55.09$\pm$8.19(8) & 44.97$\pm$5.43(12) & 69.21$\pm$3.46(2) & 29.64$\pm$5.04(19) \\
				wdbc & 69.98$\pm$3.65(19) [-26.80\%B, -2.88\%A] & 65.13$\pm$0.08(20) & 93.82$\pm$8.30(2) & 90.47$\pm$7.91(5) & 82.03$\pm$3.28(13) & 95.60$\pm$6.12(1) & 93.64$\pm$3.06(4) & 89.08$\pm$6.97(6) & 84.30$\pm$4.44(9) & 87.44$\pm$5.59(7) \\
				letter & 67.90$\pm$5.78(2) [-17.25\%B, +429.32\%A] & 82.06$\pm$0.05(1) & 9.33$\pm$0.00(11) & 8.57$\pm$0.13(18) & 8.70$\pm$0.00(17) & 12.80$\pm$1.17(5) & 11.26$\pm$0.00(7) & 8.93$\pm$0.05(13.5) & 9.53$\pm$0.51(10) & 8.26$\pm$0.00(21) \\
				ionosphere & 96.34$\pm$1.16(12) [-2.74\%B, +5.69\%A] & 98.41$\pm$0.00(2) & 97.75$\pm$0.76(6) & 98.22$\pm$1.02(3) & 97.95$\pm$0.69(5) & 99.06$\pm$0.32(1) & 94.58$\pm$1.57(16) & 95.38$\pm$0.51(13.5) & 96.43$\pm$0.50(11) & 97.45$\pm$0.53(8) \\
				wpbc & 38.40$\pm$2.32(22) [-57.01\%B, -23.19\%A] & 47.16$\pm$0.03(8) & 67.36$\pm$4.34(6) & 69.02$\pm$13.91(5) & 46.11$\pm$2.74(10) & 89.31$\pm$5.37(1) & 41.20$\pm$2.62(15) & 87.45$\pm$6.21(2) & 54.61$\pm$3.75(7) & 40.88$\pm$3.04(17) \\
				satellite & 86.73$\pm$0.41(5) [-6.28\%B, +6.85\%A] & 92.54$\pm$0.00(1) & 88.05$\pm$0.00(3) & 85.83$\pm$0.72(9) & 86.01$\pm$0.00(7) & 87.62$\pm$0.24(4) & 85.86$\pm$0.00(8) & 88.64$\pm$0.07(2) & 85.09$\pm$0.18(11) & 80.90$\pm$0.00(16) \\
				satimage-2 & 93.41$\pm$2.80(11) [-4.98\%B, +10.21\%A] & 93.29$\pm$0.03(12) & 92.78$\pm$0.00(14) & 96.16$\pm$0.00(5) & 96.69$\pm$0.00(4) & 94.70$\pm$1.19(8) & 88.46$\pm$0.00(17) & 95.44$\pm$0.23(7) & 88.05$\pm$5.10(18) & 96.92$\pm$0.00(2) \\
				landsat & 55.01$\pm$0.34(6) [-15.69\%B, +24.30\%A] & 65.24$\pm$0.01(1) & 58.24$\pm$0.00(5) & 54.52$\pm$4.05(8) & 54.85$\pm$0.00(7) & 53.12$\pm$1.57(9) & 61.37$\pm$0.00(3) & 45.10$\pm$0.17(12) & 34.83$\pm$0.91(22) & 37.01$\pm$0.00(19) \\
				celeba & 10.50$\pm$0.51(15) [-49.89\%B, -11.30\%A] & 11.49$\pm$0.01(13) & 8.39$\pm$0.38(19) & 10.65$\pm$0.49(14) & 11.92$\pm$0.50(11) & 9.74$\pm$0.68(16) & 3.61$\pm$0.15(26) & 2.90$\pm$0.92(27) & 18.03$\pm$2.60(6) & 20.27$\pm$0.90(3) \\
				spambase & 80.69$\pm$1.87(16) [-14.08\%B, +1.85\%A] & 93.91$\pm$0.00(1) & 80.99$\pm$0.00(15) & 83.65$\pm$0.58(6) & 83.32$\pm$0.00(8) & 86.78$\pm$0.59(3) & 72.71$\pm$0.00(24) & 85.64$\pm$0.14(4) & 72.89$\pm$0.42(23) & 82.19$\pm$0.00(9) \\
				campaign & 46.22$\pm$1.73(15) [-13.65\%B, +9.75\%A] & 53.53$\pm$0.01(1) & 34.24$\pm$0.47(21) & 49.95$\pm$0.67(3) & 49.04$\pm$0.00(7) & 48.90$\pm$0.98(8) & 40.24$\pm$0.00(18) & 48.11$\pm$0.12(13) & 48.87$\pm$0.29(9) & 49.43$\pm$0.00(6) \\
				optdigits & 18.96$\pm$7.15(12) [-62.79\%B, -1.44\%A] & 39.69$\pm$0.02(5) & 31.88$\pm$0.00(7) & 31.75$\pm$4.86(8) & 29.11$\pm$0.00(9) & 50.94$\pm$8.59(1) & 43.63$\pm$0.00(2) & 36.30$\pm$0.94(6) & 25.56$\pm$4.25(10) & 6.92$\pm$0.00(18) \\
				mnist & 64.59$\pm$4.98(13) [-16.24\%B, +17.94\%A] & 77.11$\pm$0.02(1) & 57.97$\pm$0.00(16) & 73.68$\pm$1.31(2) & 72.72$\pm$0.00(3) & 68.45$\pm$1.63(6) & 70.97$\pm$0.00(4) & 68.39$\pm$0.46(7) & 62.42$\pm$4.39(14) & 66.20$\pm$0.00(9) \\
				musk & 100.00$\pm$0.00(8) [+0.00\%B, +14.43\%A] & 100.00$\pm$0.00(8) & 97.00$\pm$2.27(18) & 100.00$\pm$0.00(8) & 100.00$\pm$0.00(8) & 92.21$\pm$6.32(20) & 100.00$\pm$0.00(8) & 100.00$\pm$0.00(8) & 100.00$\pm$0.00(8) & 100.00$\pm$0.00(8) \\
				backdoor & 89.90$\pm$0.72(1) [+0.81\%B, +197.53\%A] & 89.01$\pm$0.00(3) & 43.86$\pm$14.97(10) & 45.70$\pm$12.50(9) & 46.54$\pm$1.41(8) & 89.18$\pm$1.04(2) & 53.47$\pm$2.60(6) & 4.83$\pm$0.10(27) & 14.20$\pm$0.63(15) & 7.66$\pm$0.06(21) \\
				speech & 4.79$\pm$0.46(2) [-4.74\%B, +51.71\%A] & 5.03$\pm$0.01(1) & 2.94$\pm$0.31(17) & 3.17$\pm$0.00(11) & 2.80$\pm$0.00(21) & 3.38$\pm$0.50(6.5) & 3.15$\pm$0.00(12) & 3.10$\pm$0.06(13) & 3.00$\pm$0.29(14) & 2.78$\pm$0.00(23) \\
				census & 18.88$\pm$0.88(11) [-34.86\%B, +12.87\%A] & 21.71$\pm$0.01(2) & 16.01$\pm$2.97(13) & 21.05$\pm$0.67(5) & 21.68$\pm$0.63(3) & 21.20$\pm$0.61(4) & 13.71$\pm$0.42(21) & 14.96$\pm$4.41(15) & 19.67$\pm$0.60(10) & 20.32$\pm$0.66(7) \\
				CIFAR10 & 16.84$\pm$2.25(17) [-24.13\%B, -2.48\%A] & 21.01$\pm$0.00(3) & 17.45$\pm$0.52(14) & 19.91$\pm$0.00(6) & 19.62$\pm$0.00(8) & 17.39$\pm$0.59(15) & 22.17$\pm$0.00(2) & 19.98$\pm$0.07(5) & 19.55$\pm$0.08(9) & 19.42$\pm$0.00(10) \\
				SVHN & 14.46$\pm$1.53(16) [-12.78\%B, +4.27\%A] & 16.58$\pm$0.00(1) & 14.67$\pm$0.33(15) & 15.53$\pm$0.00(5) & 15.34$\pm$0.00(7) & 15.60$\pm$0.31(4) & 15.97$\pm$0.00(3) & 15.40$\pm$0.26(6) & 15.08$\pm$0.04(9) & 15.00$\pm$0.00(11) \\
				MVTec-AD & 73.05$\pm$14.04(14) [-18.35\%B, +2.63\%A] & 78.66$\pm$0.01(6) & 73.25$\pm$1.79(13) & 82.94$\pm$2.68(4) & 75.76$\pm$2.71(9) & 89.46$\pm$2.23(1) & 75.79$\pm$2.95(7) & 87.91$\pm$2.31(2) & 73.66$\pm$2.72(12) & 73.03$\pm$2.82(15) \\
				FashionMNIST & 59.30$\pm$20.52(8) [-7.26\%B, +25.20\%A] & 59.81$\pm$0.01(4) & 52.19$\pm$2.78(16) & 59.79$\pm$0.00(5) & 59.15$\pm$0.00(9) & 63.08$\pm$0.96(3) & 63.61$\pm$0.00(2) & 59.60$\pm$0.17(7) & 57.14$\pm$0.12(11) & 56.53$\pm$0.00(13) \\
				MNIST-C & 56.98$\pm$22.01(1) [+9.26\%B, +58.23\%A] & 51.24$\pm$0.00(5) & 38.40$\pm$3.34(16) & 47.42$\pm$0.00(6) & 46.20$\pm$0.00(8) & 51.47$\pm$1.10(4) & 51.89$\pm$0.00(3) & 46.89$\pm$0.14(7) & 41.78$\pm$0.12(11) & 41.57$\pm$0.00(12) \\
				yelp & 15.94$\pm$1.27(6) [-12.15\%B, +24.40\%A] & 18.14$\pm$0.00(1) & 13.90$\pm$3.00(7) & 16.35$\pm$0.00(2) & 16.03$\pm$0.00(5) & 10.40$\pm$0.09(22) & 16.14$\pm$0.00(3) & 9.96$\pm$0.05(25) & 12.80$\pm$0.02(15) & 13.42$\pm$0.00(10) \\
				imdb & 9.40$\pm$0.25(10) [-51.27\%B, -2.35\%A] & 9.23$\pm$0.00(12) & 10.53$\pm$2.36(2) & 8.97$\pm$0.00(17.5) & 8.92$\pm$0.00(20) & 10.24$\pm$0.13(4) & 9.03$\pm$0.00(14.5) & 9.79$\pm$0.03(6) & 8.70$\pm$0.02(26) & 8.85$\pm$0.00(21) \\
				agnews & 26.03$\pm$5.84(1) [+0.48\%B, +86.84\%A] & 24.88$\pm$0.00(4) & 15.16$\pm$3.17(8) & 17.35$\pm$0.00(5) & 16.68$\pm$0.00(6) & 15.42$\pm$0.38(7) & 25.86$\pm$0.00(3) & 12.30$\pm$0.06(14) & 11.85$\pm$0.03(17) & 12.82$\pm$0.00(11) \\
				20news & 17.02$\pm$8.08(2) [-8.93\%B, +33.53\%A] & 18.69$\pm$0.00(1) & 13.51$\pm$0.84(10) & 15.61$\pm$1.35(3) & 13.47$\pm$0.52(11) & 14.62$\pm$0.88(7) & 15.04$\pm$0.67(5) & 13.59$\pm$0.34(9) & 11.48$\pm$0.44(19) & 11.83$\pm$0.52(15) \\
				amazon & 12.01$\pm$0.42(2) [-6.43\%B, +11.83\%A] & 12.84$\pm$0.00(1) & 11.10$\pm$1.46(8.5) & 11.73$\pm$0.00(3) & 11.69$\pm$0.00(5) & 10.19$\pm$0.08(20) & 11.04$\pm$0.00(13) & 9.74$\pm$0.02(24) & 10.76$\pm$0.02(15) & 11.06$\pm$0.00(11.5) \\
				internetads & 89.06$\pm$0.42(1) [+5.35\%B, +86.36\%A] & 84.54$\pm$0.02(2) & 38.29$\pm$5.22(22) & 51.30$\pm$2.02(9) & 49.22$\pm$0.00(12) & 60.03$\pm$1.39(6) & 50.43$\pm$0.00(10) & 60.52$\pm$1.05(5) & 47.70$\pm$0.16(14) & 48.15$\pm$0.00(13) \\
				\bottomrule
			\end{tabular}%
		}
	\end{table*}
	
	Table \ref{tab:wilcoxon_pvalues} reports Wilcoxon signed-rank test results comparing uLEAD-TabPFN with each of the top-10 baseline methods under ROC-AUC and PR-AUC. Tests are conducted separately for low-, medium-, and high-dimensional dataset groups, as well as across all datasets. The ``Wins'' column summarizes how often uLEAD-TabPFN significantly outperforms competing methods. Consistent with the main results, statistically significant advantages are most frequent in the high-dimensional setting, supporting the conclusion that uLEAD-TabPFN is particularly effective when feature dimensionality and dependency complexity increase.
	
	\begin{table*}[htbp]
		\centering
		\caption{Wilcoxon signed-rank test $p$-values comparing \textbf{uLEAD-TabPFN} against the top-10 baseline methods for ROC-AUC and PR-AUC, grouped by feature dimensionality. The ``Wins'' column reports the number of datasets where uLEAD-TabPFN is significantly better than baselines (over all baselines / within the top-10 set). Statistical significance is indicated by $^{*}p<0.05$, $^{**}p<0.01$, and $^{***}p<0.001$.}
		\label{tab:wilcoxon_pvalues}
		\resizebox{0.9\textwidth}{!}{
			\begin{tabular}{l | c | c c c c c c c c c}
				\toprule
				\textbf{ROC-AUC} & \textbf{Wins} & FoMo-0D & DTE-NP & kNN & ICL & LOF & CBLOF & FeatBag & SLAD & DDPM \\
				\midrule
				$p$-value ($d\leq20$) & 6/25 & 0.9677 & 0.9865 & 0.9876 & 0.7634 & 0.8202 & 0.8962 & 0.2031 & 0.3420 & 0.8739 \\
				$p$-value ($20<d\leq100$) & 11/25 & 0.5675 & 0.7525 & 0.7387 & 0.4493 & 0.1519 & 0.3830 & 0.1061 & 0.2754 & 0.0594 \\
				$p$-value ($d>100$) & 20/25 & 0.0004$^{***}$ & 0.2352 & 0.0886 & 0.1384 & 0.2163 & 0.0207$^{*}$ & 0.2352 & 0.0320$^{*}$ & 0.0038$^{**}$ \\
				$p$-value (All) & 17/25 & 0.3951 & 0.9477 & 0.9233 & 0.4573 & 0.3328 & 0.3123 & 0.0464$^{*}$ & 0.0357$^{*}$ & 0.1137 \\
				\midrule\midrule
				
				\textbf{PR-AUC} & \textbf{Wins} & DDAE & FoMo-0D & DTE-NP & kNN & ICL & LOF & SLAD & DDPM & OCSVM \\
				\midrule
				$p$-value ($d\leq20$) & 9/26 & 0.9997 & 0.9830 & 0.9903 & 0.9911 & 0.6580 & 0.6371 & 0.5279 & 0.8739 & 0.6051 \\
				$p$-value ($20<d\leq100$) & 13/26 & 0.9940 & 0.3830 & 0.7789 & 0.7659 & 0.4831 & 0.3047 & 0.3353 & 0.1733 & 0.2475 \\
				$p$-value ($d>100$) & 20/26 & 0.8187 & 0.0017$^{**}$ & 0.2163 & 0.0789 & 0.1147 & 0.1501 & 0.0368$^{*}$ & 0.0093$^{**}$ & 0.0093$^{**}$ \\
				$p$-value (All) & 16/26 & 1.0000 & 0.6019 & 0.9630 & 0.9432 & 0.3890 & 0.3447 & 0.1805 & 0.3123 & 0.1783 \\
				\bottomrule
			\end{tabular}
		}
	\end{table*}
			
	\section{Appendix: A Detailed  Case Study}
	\label{app:case}
		
	We present an end-to-end case study on the \textit{Wilt} dataset to illustrate how uLEAD-TabPFN operates in practice, from latent representation learning to dependency-based scoring and interpretation. Wilt is a widely used benchmark for tabular anomaly detection derived from high-resolution remote sensing imagery, where diseased trees correspond to anomalous instances. The dataset consists of five continuous spectral and texture-related features extracted from multispectral and panchromatic image bands, including mean values of the green (Mean-G), red (Mean-R), and near-infrared (Mean-NIR) bands, as well as texture statistics from the panchromatic band (GLCM-Pan) and intensity variation measured by the standard deviation of the panchromatic band (SD-Pan). With a total of 4,819 samples (4,562 normal and 257 anomalous), its low dimensionality and well-defined physical feature semantics make Wilt particularly suitable for a detailed methodological analysis and interpretability study. A complete and detailed walk-through of this case study is provided in Appendix~\ref{app:case}.
	
	On the Wilt dataset, uLEAD-TabPFN achieves a ROC-AUC of 97.5\% and a PR-AUC of 90.8\%, substantially outperforming existing baseline methods, which typically achieve ROC-AUC values around 60\% and PR-AUC values near 20\%. Relative to the average baseline performance, this corresponds to improvements of 81\% in ROC-AUC and 534\% in PR-AUC.
	
	The goal of this case study is to understand why uLEAD-TabPFN performs particularly well on this dataset. To this end, we compare uLEAD-TabPFN with the ablated variants introduced in Section~\ref{sec:ablation}. Specifically, uLEAD-Original, which operates directly on the original feature space without latent transformation, achieves 84\% ROC-AUC and 43\% PR-AUC. Replacing TabPFN with a CART-based dependency estimator (uLEAD-CART) obtains performance to 86\% ROC-AUC and 51\% PR-AUC. LEAD-TabPFN without uncertainty-aware scoring attains the same ROC-AUC (97.5\%) and PR-AUC (90.8\%) as the full model.
	
	These results indicate that the dominant performance gains on the Wilt dataset stem from the representation learning and dependency modeling enabled by TabPFN, while uncertainty-aware scoring plays a less critical role in this particular setting. Consequently, the following analysis focuses on understanding how the latent space transformation contributes to improved anomaly detection performance.
	
	\paragraph{Latent Space Analysis.}
	\begin{table}[htbp]
		\centering
		\caption{Comparison of geometric and neighborhood properties in the original and latent spaces on the Wilt dataset.}
		\label{tab:latent_distance_analysis}
		%\small
		\resizebox{\columnwidth}{!}{%
			\begin{tabular}{p{3cm}ccp{5cm}}
				\toprule
				\textbf{Metric} & \textbf{Original} & \textbf{Latent} & \textbf{Interpretation} \\
				\midrule
				Average silhouette score (all samples) 
				& $-0.1402$ & $0.0024$ 
				& Normal and anomalous samples are highly overlapping in both spaces, with slightly improved separation in the latent space. \\
				
				Average distance to nearest normal sample 
				& $0.3316$ & $0.3387$ 
				& The separation between normal and anomalous samples remain similar across spaces. \\
				
				LOF ratio (anomaly / normal) 
				& $1.0383$ & $1.0297$ 
				& Density-based contrast remains limited and slightly weaker in the latent space. \\
				
				Anomalies with anomaly as nearest neighbor
				& $57.2\%$ & $71.2\%$ 
				& Anomalies exhibit stronger local neighborhood coherence in the latent space. \\
				
				30-NN purity (fraction of anomalies) 
				& $0.2943$ & $0.3926$ 
				& Anomalies form more locally coherent neighborhoods in the latent space. \\
				\bottomrule
			\end{tabular}
		}
	\end{table}
	We first compare feature dependencies in the original and latent spaces using the Pearson correlation coefficient and the conditional coefficient of determination ($R^2$) estimated with TabPFN. Pearson correlation measures pairwise linear dependence, while conditional $R^2$ quantifies how well each feature can be predicted from the remaining features.
	
	In the original feature space, the mean absolute Pearson correlation is 0.239, indicating weak pairwise dependencies. After projection into the latent space, this value increases to 0.531 (a 122.4\% relative increase), with the standard deviation rising from 0.317 to 0.581, indicating amplified and more diverse dependency strengths. Consistent with this, the average conditional $R^2$ increases from 0.649 (std. 0.285) in the original space to 0.991 (std. 0.003) in the latent space, which shows that the learned representation makes conditional relationships substantially more predictable and stable in the latent space.
		
\begin{figure*}
		\centering
		\includegraphics[width=0.9\linewidth]{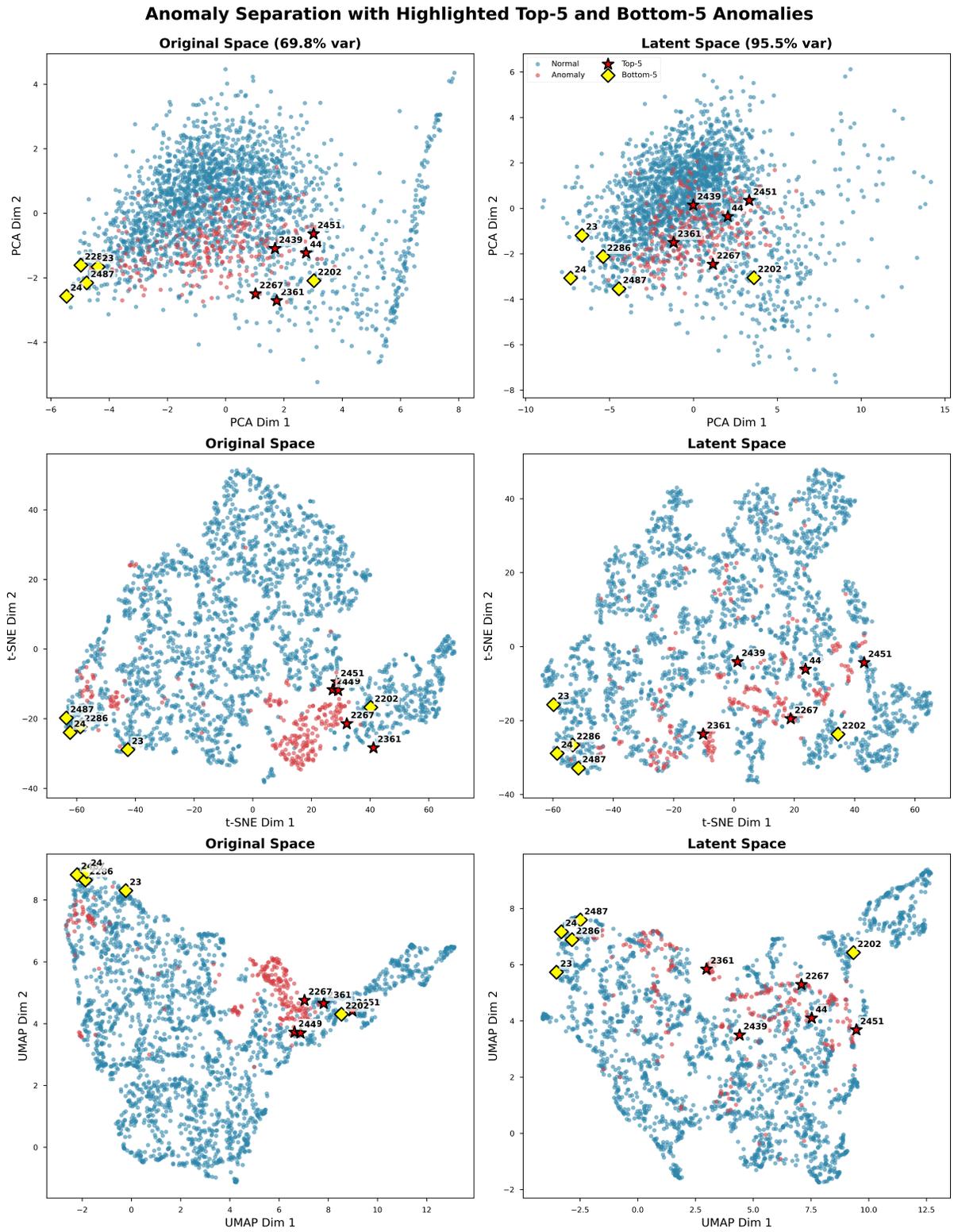}
		\caption{Visulaization of top-5 and bottom-5 anomalies in both the original and latent feature spaces.}
		\label{fig:top_anomalies}
	\end{figure*}

	To examine how normal and anomalous samples are distributed in the original and latent spaces, we compare the two representations using multiple dimensionality reduction techniques, including PCA, t-SNE, and UMAP as shown in Figure~\ref{fig:top_anomalies}. 
		
	In addition, Table~\ref{tab:latent_distance_analysis} reports complementary quantitative metrics characterizing geometric separation, density contrast, and neighborhood structure in both spaces. Specifically, the silhouette score measures the degree of separation between normal and anomalous samples based on relative intra- and inter-class distances, with values near zero indicating substantial overlap. The distance to the nearest normal sample and the Local Outlier Factor (LOF) ratio capture marginal distance- and density-based distinguishability, while nearest-neighbor anomaly rates and $k$-NN purity quantify local neighborhood coherence among anomalous samples.
	
	Figure~\ref{fig:wilt_umap} shows that normal and anomalous samples remain substantially overlapping in both the original and latent spaces, consistent with the near-zero silhouette scores reported in Table~\ref{tab:latent_distance_analysis} (from $-0.140$ in the original space to $0.002$ in the latent space). Likewise, the average distance from anomalous samples to their nearest normal neighbors changes only marginally, from $0.332$ to $0.339$, and density-based contrast remains weak, with the LOF ratio slightly decreasing from $1.038$ to $1.030$. Together, these results indicate that the improved detection performance of uLEAD-TabPFN is not driven by global geometric separation or marginal distance- and density-based cues.
	
	In contrast, both the visualization and neighborhood-level metrics reveal a clear structural change in the latent space. As reported in Table~\ref{tab:latent_distance_analysis}, the proportion of anomalies whose nearest neighbor is also anomalous increases from $57\%$ to $71\%$, and the fraction of anomalies among the 30 nearest neighbors rises from $0.29$ to $0.39$, indicating stronger local neighborhood coherence among anomalous samples. In addition, the average silhouette score computed over anomalous samples remains positive in both spaces ($0.331$ in the original space and $0.275$ in the latent space), whereas the corresponding scores for normal samples are negative ($-0.167$ and $-0.013$). Together, these results indicate that anomalous samples exhibit more coherent local neighborhood structure than normal samples in both representations, with stronger neighborhood-level coherence in the latent space despite slightly reduced global compactness.
	
	All these indicate that in the original feature space, anomalies tend to form a single broad cluster that substantially overlaps with normal samples, which limit the effectiveness of distance-, density-, and marginal-distribution-based baseline methods. In contrast, Figure~\ref{fig:wilt_umap} illustrates that anomalous samples in the latent space are organized into multiple locally coherent neighborhoods rather than a single globally compact cluster. Although anomalies are not more geometrically separated from normal samples, the latent representation restructures anomalous instances according to shared dependency violations. This organization aligns naturally with the dependency-based scoring of uLEAD-TabPFN, where anomaly detection is driven by conditional inconsistency rather than marginal distance or density.

	\paragraph{Anomaly Interpretation.}
	uLEAD-TabPFN provides interpretability by design through its explicit dependency-based anomaly scoring mechanism. For a given test sample, the anomaly score is computed as an aggregation of normalized conditional dependency deviations across latent dimensions. Latent dimensions with large contributions correspond to conditional relationships that are unlikely under the learned normal data distribution, making the source of abnormality directly observable at the representation level.
	
	Crucially, the latent representation is obtained via a linear transformation of the original feature space. As a result, dependency deviations detected in the latent space can be directly traced back to the original input features through the encoder weights, enabling faithful feature-level attribution without relying on post-hoc explanation techniques. This interpretability reflects conditional inconsistency rather than marginal outliers: a feature may appear anomalous not because of its absolute value, but because it violates learned dependencies given the remaining features. This aligns the explanation with the underlying detection mechanism of uLEAD-TabPFN.
	
	We analyze a representative top-ranked anomaly from the Wilt dataset to illustrate instance-level interpretability. For this sample, the anomaly score is primarily driven by four out of five latent dimensions. Specifically, latent dimensions $Z_0$, $Z_1$, $Z_2$, and $Z_4$ contribute $41.5$, $43.8$, $39.3$, and $32.9$ to the total anomaly score, respectively, while $Z_3$ contributes only $7$. 
	
	\begin{figure}
		\centering
		\includegraphics[width=\columnwidth]{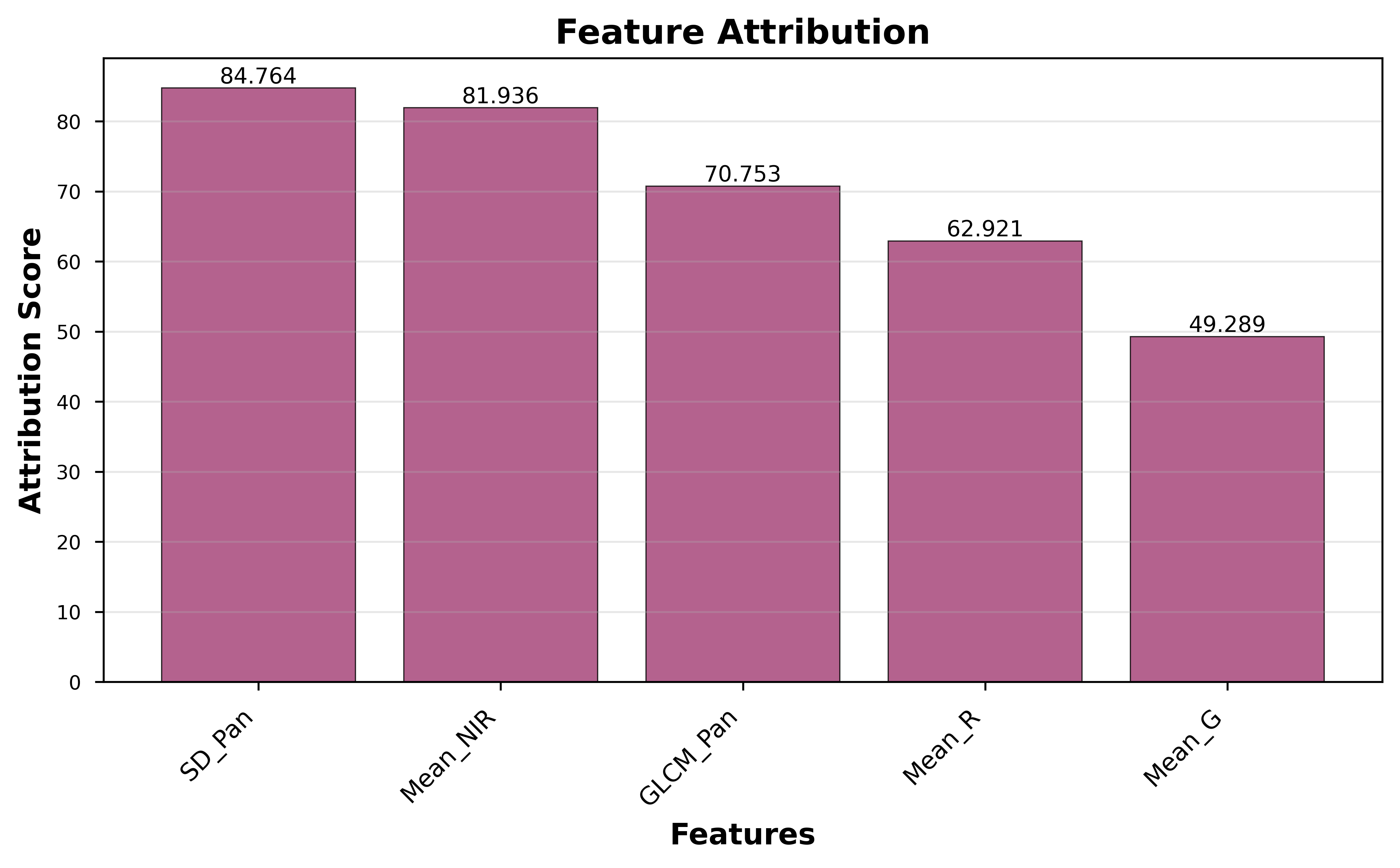}
		\caption{The attribution of the orginal features to the anomaly}
		\label{fig:anomaly_attribution}
	\end{figure}
	
	Figure~\ref{fig:anomaly_attribution} visualizes how these latent-space dependency violations are attributed back to the original input features. Owing to the linear encoder, the contribution of each original feature to the anomaly score can be explicitly quantified. For this anomaly, the strongest attributions correspond to features associated with vegetation condition, such as near-infrared reflectance (Mean-NIR) and panchromatic intensity variation (SD-Pan). Similar attribution patterns are consistently observed across other top-ranked anomalies, as documented in the appendix.
	
	Overall, this case study demonstrates how uLEAD-TabPFN enables transparent, end-to-end interpretation by linking latent dependency violations, uncertainty-aware scoring, and original feature contributions within a unified framework. A complete walk-through on the Wilt dataset, including detailed per-dimension statistics, encoder weights, and additional anomaly analyses.
	
	\begin{table}[htbp]
		\centering
		\caption{Original feature values for anomaly \#1 showing raw values, normalized values, and attribution scores across all features}
		\resizebox{1\columnwidth}{!}{
			\begin{tabular}{lrrrrr}
				\toprule
				Metric & GLCM\_Pan & Mean\_G & Mean\_R & Mean\_NIR & SD\_Pan \\
				\midrule
				Raw Value & 150.181 & 225.000 & 145.682 & 540.500 & 13.908 \\
				Raw Percentile & 92.4\% & 56.8\% & 86.7\% & 52.8\% & 14.7\% \\
				Normalized Value & 2.499 & 0.269 & 3.714 & 0.129 & -1.509 \\
				Attribution Score & 5.181 & 3.649 & 3.978 & 6.193 & 5.240 \\
				\bottomrule
			\end{tabular}
		}
		\label{tab:anomly1_values}
	\end{table}
	
	\begin{table}[htbp]
		\centering
		\caption{Latent feature analysis for anomaly \#1 showing latent values, predicted means, deviations, variances, negative log-likelihoods, and contributions for each latent dimension}
		\resizebox{1\columnwidth}{!}{
			\begin{tabular}{lrrrrr}
				\toprule
				Metric & $Z_0$ & $Z_1$ & $Z_2$ & $Z_3$ & $Z_4$ \\
				\midrule
				Latent Value & -1.352 & -0.627 & -1.373 & -0.024 & 0.247 \\
				Latent Percentile & 12.9\% & 26.6\% & 19.4\% & 61.4\% & 54.6\% \\
				Conditional Mean & -0.284 & -2.146 & 0.569 & 0.316 & 2.006 \\
				Residual & 1.068 & 1.519 & 1.942 & 0.340 & 1.760 \\
				Conditional Variance & 1.000 & 1.000 & 1.000 & 1.000 & 1.000 \\
				NLL & 1.489 & 2.072 & 2.805 & 0.977 & 2.467 \\
				Contribution & 1.140 & 2.306 & 3.771 & 0.116 & 3.096 \\
				\bottomrule
			\end{tabular}
		}
		\label{tab:anomly1_latent}
	\end{table}
		
	We analyze the interpretability of the highest-ranked anomaly (Index: 2361, Score: 75.3610).
	Original feature values for anomaly \#1, including raw values, normalized values, and attribution scores across all features, are reported in Table~\ref{tab:anomly1_values}.
	
	Latent feature analysis for anomaly \#1, including latent values, predicted means, deviations, variances, negative log-likelihoods, and per-dimension contributions, is reported in Table~\ref{tab:anomly1_latent}.

\end{document}